\documentclass[conference]{IEEEtran}
\IEEEoverridecommandlockouts
\usepackage{amsmath,amsfonts}
\usepackage{algorithmic}
\usepackage{algorithm}
\usepackage{array}
\usepackage{textcomp}
\usepackage{stfloats}
\usepackage{url}
\usepackage{verbatim}
\usepackage{graphicx}
\usepackage{cite}
\usepackage[hidelinks]{hyperref}
\usepackage{color,soul}
\usepackage{amssymb}
\usepackage{tabularx}
\usepackage{caption}
\usepackage{subcaption}
\usepackage{amsmath}
\usepackage{xcolor}
\usepackage{booktabs}
\usepackage{multirow}

\usepackage{flushend}

\def\BibTeX{{\rm B\kern-.05em{\sc i\kern-.025em b}\kern-.08em
		T\kern-.1667em\lower.7ex\hbox{E}\kern-.125emX}}

\begin{document}
	
	\title{Too Good To Be True: performance overestimation in (re)current practices for Human Activity Recognition}
	
	\author{
		% \IEEEauthorblockN{anonymous authors}
		 \IEEEauthorblockN{Andr\'{e}s Tello}
		 \IEEEauthorblockA{\textit{Bernoulli Institute} \\
			 	\textit{University of Groningen}\\
			 	Groningen, The Netherlands \\
			 	andres.tello@rug.nl}
		 \and
		 \IEEEauthorblockN{Victoria Degeler}
		 \IEEEauthorblockA{\textit{Informatics Institute} \\
			 	\textit{University of Amsterdam}\\
			 	Amsterdam, The Netherlands \\
			 	v.o.degeler@uva.nl}
		 \and
		 \IEEEauthorblockN{Alexander Lazovik}
		 \IEEEauthorblockA{\textit{Bernoulli Institute} \\
			 	\textit{University of Groningen}\\
			 	Groningen, The Netherlands \\
			 	a.lazovik@rug.nl}	
	}
	
	\maketitle

	\begin{abstract}
		%Human Activity Recognition has been attracting the interest of academia and industry for several decades.
		Today, there are standard and well established procedures within the Human Activity Recognition (HAR) pipeline. However, some of these conventional approaches lead to accuracy overestimation. In particular, sliding windows for data segmentation followed by standard random k-fold cross validation, produce biased results. An analysis of previous literature and present-day studies, surprisingly, shows that these are common approaches in state-of-the-art studies on HAR. It is important to raise awareness in the scientific community about this problem, whose negative effects are being overlooked. Otherwise, publications of biased results lead to papers that report lower accuracies, with correct unbiased methods, harder to publish. Several experiments with different types of datasets and different types of classification models allow us to exhibit the problem and show it persists independently of the method or dataset.
	\end{abstract}
	
	\begin{IEEEkeywords}
		Human Activity Recognition, HAR, Classification Models, Performance Overestimation, Biased Accuracy, Random K-Fold Cross-Validation
	\end{IEEEkeywords}
	
	\section{Introduction}
	\label{sec:intro}
	
	Human Activity Recognition is an ongoing research topic in the fields of ubiquitous and pervasive computing, healthcare, ambient assisted living, among others. Several methods has been proposed for HAR, from traditional machine learning algorithms \cite{van2008accurate, bulling2014tutorial, kwapisz2011activity}, to current Deep Learning approaches \cite{chen2015deep, hammerla2016deep, ordonez2016deep, wan2020deep}. With either approach, supervised learning is commonly used to learn models that classify activities based on annotated sensor data collected from an instrumented testing environment, e.g., a smart home, wearable IMU sensors. Generally, HAR implementations include data collection, pre-processing, data segmentation, feature extraction, and classification.
	
	Some previous studies on HAR noticed that conventional methods used within the HAR pipeline can lead to accuracy overestimation. Hammerla and Pl{\"o}tz~\cite{hammerla2015let} proved that standard k-fold cross validation (CV) are biased due to statistical dependence between data samples, specially when using sliding windows for data segmentation. The problem was also mentioned in \cite{dehghani2019subject, dehghani2019quantitative, jordao2018human, gholamiangonabadi2020deep, ahad2020iot}, although these papers focused on different goals, and did not go into analyzing the problem in detail. 
	Surprisingly, it is a widely-used ongoing practice within the HAR domain. We performed an extensive review of the recent works in the HAR community. 
	Table \ref{table:prevwork} shows a list of recent studies where \textit{sliding windows segmentation} and \textit{random K-fold CV} are used in the HAR pipeline. While not exhaustive, this list includes remarkable works from previous years, and a growing number of present-day studies. These works have been published in top journals or presented at top conferences which have created a high impact in the HAR community. 
	
	\begin{table*}[h!]
		\caption{HAR studies following a sliding window data segmentation and random training/test split validation approach.}
		\centering
		\resizebox{0.98\textwidth}{!}{%
			\begin{tabular}{l l l} 
				\toprule
				\textbf{Authors} & \textbf{Year, (Citations)} & \textbf{Journal/Conference \hfill (Impact Factor)} \\ 
				\midrule
				% Altun et al. \cite{altun2010comparative} & 2010, (623) & Pattern Recognition \hfill (8.5) \\ 
				% Reiss et al. \cite{reiss2013confidence} & 2013, (32) & UBICOMP 2013 \hfill - \\  
				% Velloso et al. \cite{velloso2013qualitative} & 2013, (158) & Conference Proceedings \hfill - \\ 
				% Bogomolov et al. \cite{bogomolov2014pervasive} & 2014, (76) & Conference Proceedings \hfill - \\
				% Banos et al. \cite{banos2014window} & 2014, (570) & Sensors \hfill (3.6) \\  
				% Krishnan et al. \cite{krishnan2014activity} & 2014, (622) & Pervasive and Mobile Computing \hfill (3.8)\\ 
				Khalifa et al. \cite{khalifa2017harke} & 2017, (148) & IEEE Transactions on Mobile Computing \hfill (6.1)\\ 
				Micucci et al. \cite{micucci2017unimib} & 2017, (464) & Applied Sciences \hfill (2.8)\\ 
				San-Segundo et al. \cite{san2018robust} & 2018, (93) & Engineering Applications of Artificial Intelligence, \hfill (7.8)\\ 
				Wang et al. \cite{wang2018eating} & 2018, (28) & Smart Health \hfill (5.1)\\ 
				%				Zhang et al. \cite{zhang2018multi} & 2018, (37) & International Joint Conference on Artificial Intelligence \hfill - \\ 
				Mutegeki and Han \cite{mutegeki2020cnn} & 2020, (225) & ICAIIC \hfill -\\ 
				Ni et al. \cite{ni2020leveraging} & 2020, (19) & Sensors \hfill (3.8) \\ 
				Gupta \cite{gupta2021deep} & 2021, (58) & International Journal of Information Management Data Insights \hfill -\\ 
				Li et al. \cite{li2021tribogait} & 2021, (6) & UBICOMP 2021 \hfill -\\ 
				Mekruksavanich and Jitpattanakul \cite{mekruksavanich2021lstm} & 2021, (165) & Sensors \hfill (3.8)\\ 
				Bouchabou et al. \cite{bouchabou2021fully} & 2021, (20) & Communications in Computer and Information Science \hfill -\\ 
				G{\'o}mez Ramos et al. \cite{ramos2021daily} & 2021, (16) & Sensors \hfill (3.8)\\ 
				Zimbelman and Keefe \cite{zimbelman2021development} & 2021, (10) & PLOS ONE \hfill - \\ 
				Yan et al. \cite{yan2022deep} & 2022, (10) & IEEE International Conference on Bioinformatics and Biomedicine \hfill - \\
				Wang et al. \cite{wang2022sensor} & 2022, (23) & IEEE Sensors Journal \hfill (4.3) \\			
				Huang et al. \cite{huang2022channel} & 2022, (34) & IEEE Transactions on Mobile Computing \hfill (6.1)\\
				Luo et al. \cite{luo2021binarized} & 2023, (17) & IEEE Transactions on Mobile Computing \hfill (6.1)\\			
				Wu et al. \cite{wu2023novel} & 2023, (5) & Knowledge-Based Systems \hfill (8.1) \\
				Garcia-Gonzalez et al. \cite{garcia2023new} & 2023, (8) & Knowledge-Based Systems \hfill (8.1) \\
				\bottomrule
		\end{tabular}}
		\label{table:prevwork}
	\end{table*}	
	
	Turning to the experimental evidence on performance overestimation, we evaluated the effect of sliding windows data segmentation and random splitting on model accuracy. In our experiments we used datasets with different data modality (CASAS \cite{cook2012casas} binary motion sensors, MHEALTH \cite{banos2014mhealthdroid} and PAMAP2 \cite{reiss2012intro} on-body inertial sensors) and classification models of different nature: Random Forest (RF), Graph Neural Networks (GNNs). The results show that independently of the type of data and the chosen model, the reported accuracy is highly overestimated following the aforementioned approach.
	
	The main contribution of this paper is two-fold: (1) It provides an extensive survey of recent papers in HAR domain that employ the flawed methodology which shows that the practice is wide-spread to this date and the importance of warning the community. (2) It provides a set of experiments where the problem is distilled and shown in its pure form. The experiments were performed on different types of HAR datasets and different types of supervised machine learning approaches, this shows that the problem persists in all cases and configurations.
	
	The remainder of this document is as follows. Section \ref{sec:problem} describes the methodological issues in conventional approaches for HAR. Section \ref{sec:evidence} presents an extensive recent related work on HAR incurring in this problem. Section \ref{sec:experiments} describes our experiments, presents the results and discussion focused on showing the effects of the problem described in Section \ref{sec:problem}. Finally, we conclude our work in Section \ref{sec:conclusions}.
	
	\section{Model performance overestimation}
	\label{sec:problem}
	
\begin{figure}[!bh]
		\centering
		\begin{subfigure}[b]{0.45\textwidth}
			\centering
			\includegraphics[width=0.75\textwidth]{"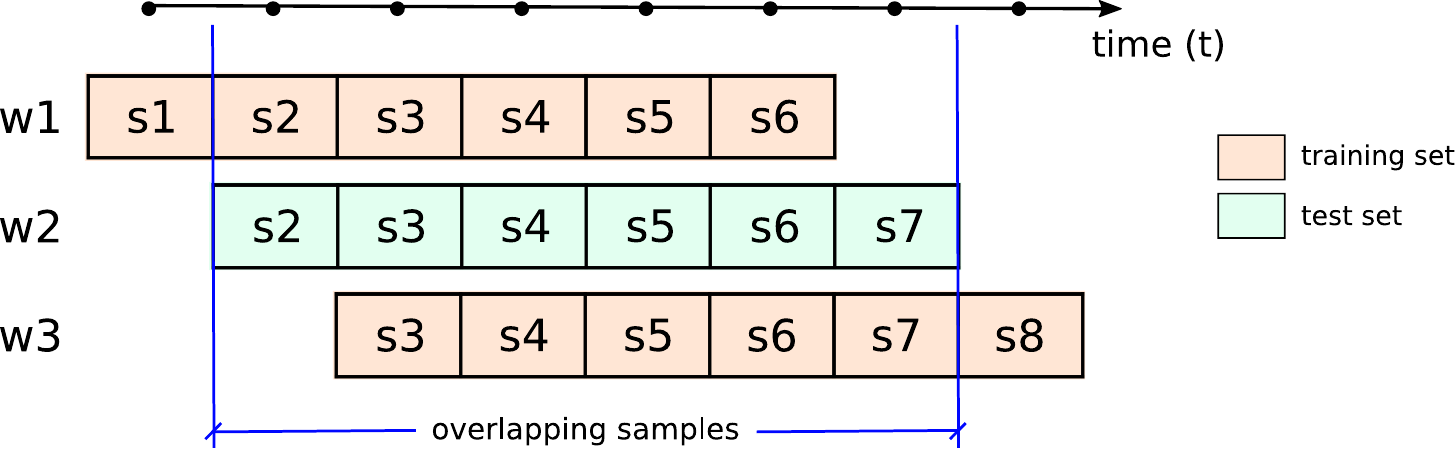"}
			\caption{overlapping sliding windows}
			\label{fig:ow}
		\end{subfigure}
		\begin{subfigure}[b]{0.45\textwidth}
			\centering
			\includegraphics[width=0.75\textwidth]{"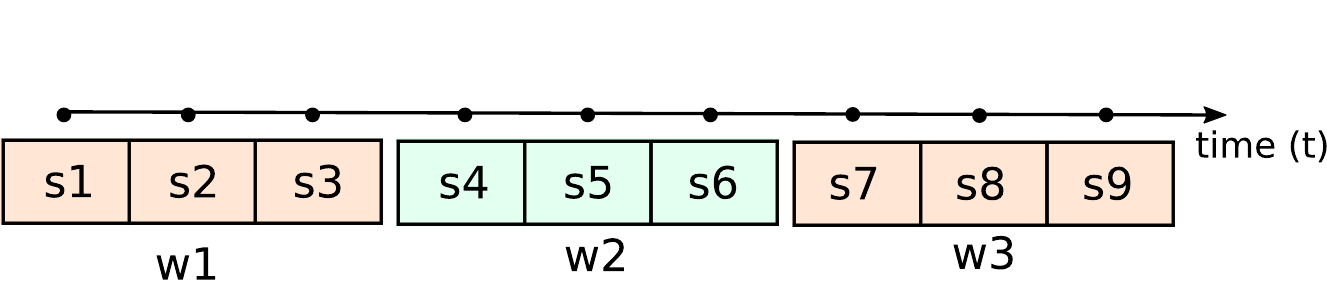"}
			\caption{non-overlapping sliding windows}
			\label{fig:now}
		\end{subfigure}
		\caption{Overlapping and non-overlapping sliding windows data segmentation}
		\label{fig:windowing}
	\end{figure}

	Data segmentation following some windowing technique is the conventional approach for HAR. Several studies proposed and evaluated different windowing mechanisms \cite{krishnan2014activity, quigley2018comparative}, the sliding windows approach being the most widely used in HAR \cite{banos2014window}. The windows can be overlapping or non-overlapping and the adopted approach may impact the performance of the classification models \cite{dehghani2019quantitative}. Figure \ref{fig:windowing} shows a window-based data segmentation. The windows can be time-based or sensor-event-based where the length is defined in \textit{\textbf{t}} seconds or \textit{\textbf{s}} number of sensor readings. The input stream of sensor readings is split into segments, i.e., windows of equal size. Then, a set of feature vectors is calculated from each window. These vectors are used as input for the classification models. 
	
	First, the models need to be trained, with the ultimate goal being to generalize to new unseen data points \cite{domingos2012few}. One way to evaluate the generalization performance of the classifier is using a hold out part of the entire dataset, i.e., the test set. The conventional approach is randomly splitting the dataset into training/test subsets. Actually, the randomness is achieved by first shuffling the entire dataset, and then splitting it  at some predefined ratio (e.g., 80:20). Other approaches split further the training set having as a result a validation subset which is used for model selection and(or) hyper-parameter optimization. In the field of HAR, the most used model evaluation technique to assess the performance of the classifiers is k-fold CV \cite{dehghani2019subject}. In this approach, the data is split into $k$ disjoint subsets of equal size. Then, the model is trained on $k$-$1$ subsets and evaluated on the $k^{th}$. This process is repeated $k$ times, taking a different subset for evaluation each time. The final performance of the model is the mean of all runs. The CV method assumption is that the data samples are Independent and Identically Distributed (i.i.d.); thereby, the order on how the samples are chosen for training/test sets does not affect the performance of the classifier \cite{hammerla2015let, dehghani2019subject}. This is where the problem on (re)current practices for HAR lies. \textit{Using sliding windows for data segmentation and feature extraction, the statistical independence assumption does not hold anymore. Therefore, random training/test split during evaluation does affect the performance of the model because of subsequent windows that are assigned one to training and the previous and(or) next to the test set}. 
	
	The problem can be observed with greater clarity in Figure \ref{fig:ow}. Windows $w1$, $w2$, and $w3$ share the samples $s3$ to $s6$. Likewise, $s2$ is shared between $w1$ and $w2$ and $s7$ is shared between $w2$ and $w3$. Therefore, the features obtained from those windows are drawn from almost the same underlying data samples. How true is then, that those new window-samples are i.i.d.? Thus, if $w2$ is randomly chosen for validation and $w1$ and $w3$ are kept for training, the classifier is tested on data that was already seen. Consequently, the performance of the classification algorithm is overestimated. 
	
	In the case of non-overlapping windows (Figure \ref{fig:now}), the dependence between consecutive windows is less evident. However, for long running activities (e.g., walking, standing, reading), the similarity between samples drawn in a short interval will create a strong correlation between consecutive windows \cite{hammerla2015let, dehghani2019subject}. Hence, it is likely that consecutive windows have similar samples and they correspond to the same activity. From Figure \ref{fig:now}, if $w2$ is randomly assigned to the test set, it will be almost identically to those assigned to the training set, i.e., $w1$ and $w3$.
	
	The problem described before creates an illusion of perfect accuracy because of overfitting, caused by the strong correlation between consecutive windows and random splitting of training and test sets. Hence, instead of learning the patterns that uniquely characterize each activity, the model will just memorize the training data and will produce the correct label if the same data is seen during testing.
	
	Performance overestimation can be avoided ensuring the independence between training and test sets. One option is applying Leave-One-Subject-Out CV (LOSO-CV), a variant of k-fold CV \cite{hammerla2015let, dehghani2019subject}. In LOSO-CV, instead of randomly choosing the samples to include in each fold, the data samples belonging to one subject are used for testing, while the data from the remaining subjects are used for training. This is repeated for each subject in the experiment. This model evaluation approach is more rigid than traditional k-fold CV and helps to avoid accuracy overestimates due to the independence between training and test sets \cite{gholamiangonabadi2020deep}. However, a LOSO-CV is not always feasible, e.g., if the number of subjects in the experiments is either too small or too large \cite{hammerla2015let}. With few users, the variability on how the activities are performed could produce an unrealistic view of the model performance. On the contrary, with too many subjects, the computational complexity during training or testing could be too high or even impractical \cite{hammerla2015let}. An alternative is to use a generalization of LOSO, the Group K-Fold CV\footnote{\scriptsize Group K-Fold CV: \href{https://scikit-learn.org/stable/modules/cross\_validation.html\#cross-validation-iterators-for-grouped-data}{https://scikit-learn.org/stable/modules/cross\_validation.html\#cross-validation-iterators-for-grouped-data}}. In this case, the data samples are grouped by a third-party parameter which can be defined per use-case basis, e.g., the collection date, subject-id, etc. Grouped partitions ensure that data samples corresponding to the same group are not represented in both testing and training sets. Hence, this group-based approach is a valid and straightforward approach to ensure an unbiased evaluation strategy.
	
	\section{(Re)current practices in HAR}
	\label{sec:evidence}
	
	The methodological issues that lead to accuracy overestimation are not only a problem of the early years of HAR research. Surprisingly, it is an ongoing practice because of the well established standard procedures, i.e., \textit{sliding windows data segmentation} and \textit{random K-fold CV}.
	
	In this section, we present a systematic review of recent scientific papers that use methodology with performance overestimation. The commonality of these works is the use of \textit{sliding windows} for data segmentation and \textit{k-fold CV} for model evaluation. This combination without unconstrained training/test splits leads to performance overestimation as described in Section \ref{sec:problem}. In some of these studies, LOSO evaluation is applied in addition to k-fold CV~\cite{micucci2017unimib, san2018robust, wang2018eating}. The results with LOSO evaluation show a significant drop in performance. However, according to the authors, the performance drop is attributed solely to the variability on the way that different subjects perform the same activities. This shows that the bias because of statistical dependence between consecutive windows and random training/test splits is overlooked and the problem described before is an ongoing practice. 
	
	%%%%% POSIBLY EXCLUDE
	Khalifa et al. \cite{khalifa2017harke} evaluated Decision Trees (DT), K-Nearest Neighbours (kNN), Naive Bayes (NB), and Support Vector Machine (SVM) models for HAR. They used the UCI Daily and Sport Activities dataset \cite{altun2010comparative} which contains 3-axis acceleration data, and their own dataset collected using a Kinetic Energy Harvesting (KEH) device built in their lab. The data from both datasets were segmented using a \textit{five-seconds sliding window with 50\% overlap}, and their models were evaluated using a \textit{10-fold CV}. The reported classification accuracy is over 86\% for the KEH signal dataset and over 97\% for the accelerometer data. The overestimated accuracy can be observed if we compare these results against those presented by Altun et al., in \cite{altun2010comparative}, who provided the results for biased (Random K-Fold) and unbiased methods (LOSO) for the same dataset. 
	
	\begin{figure}[!th]
		\includegraphics[width=0.75\columnwidth]{"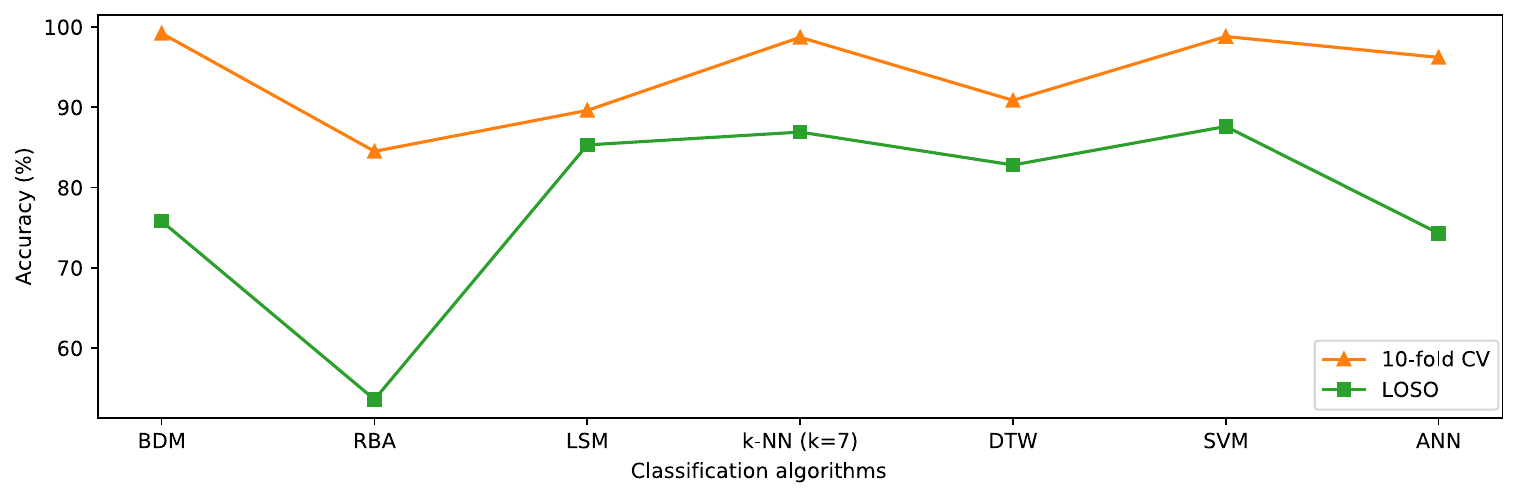"}
		\centering
		\caption{Reported classification accuracy from Altun et al., \cite{altun2010comparative} on UCI Daily and Sport Activities dataset.}
		\label{fig:altun2010}
	\end{figure}
	
\begin{figure}
		\centering
		\begin{subfigure}[b]{0.5\textwidth}
			\centering
			\includegraphics[width=0.8\textwidth]{"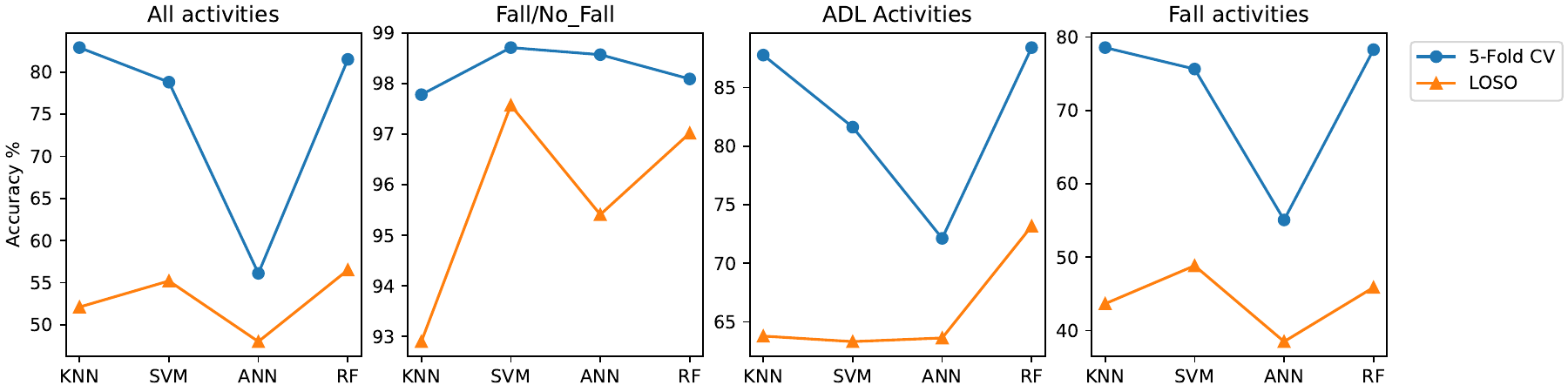"}
			\caption{Raw signal feature vector}
			\label{fig:micucci2017a}
		\end{subfigure}
		
		\begin{subfigure}[b]{0.5\textwidth}
			\centering
			\includegraphics[width=0.8\textwidth]{"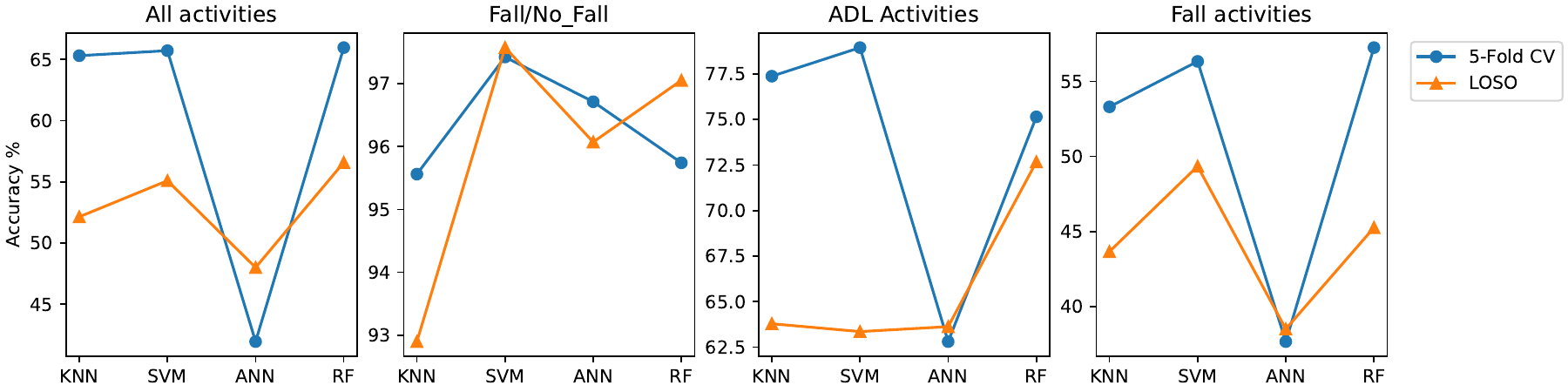"}
			\caption{Magnitude feature vector}
			\label{fig:micucci2017b}
		\end{subfigure}
		\caption{5-fold CV vs LOSO: reported accuracy comparison from Micucci et al., \cite{micucci2017unimib}}
		\label{fig:micucci2017}
	\end{figure}

	Micucci et al. \cite{micucci2017unimib} present the UniMiB-SHAR dataset, a HAR benchmark which contains smartphone acceleration data. The data were gathered from 30 subjects performing 9 types of daily living activities (ADL) and 8 different types of falls. The dataset was segmented using a fixed-size window of three seconds each time a peak was found, and as stated by the authors, this segmentation produces an \textit{overlap between subsequent windows}. They evaluated kNN, SVM, Artificial Neural Networks (ANN), and Random Forest (RF) classifiers and run four different experiments with each classifier. Each experiment used raw signals and the magnitude of the vectors as input features. The evaluation was performed using a random 5-fold CV, and LOSO-CV. The results, once again, show that 5-fold CV outperforms LOSO in all the reported experiments when using raw data features (Figure \ref{fig:micucci2017a}). Likewise, the accuracy is higher for 5-fold CV for 11 out of 16 experiments (Figure \ref{fig:micucci2017b}) using the magnitude of the signal feature vector. It is worth noting that Micucci et al. attributed the divergence of the evaluation results only to the difference in the way that different subjects perform the same activities. Moreover, they remark that a subject-dependent approach is positive because it increases the performance of the classification algorithms irrespective to the features used in the model. From these claims, it is clear that the bias introduced by overlapping windows and random training/test splits is not perceived.
	
\begin{figure}[!th]
		\includegraphics[width=0.95\columnwidth]{"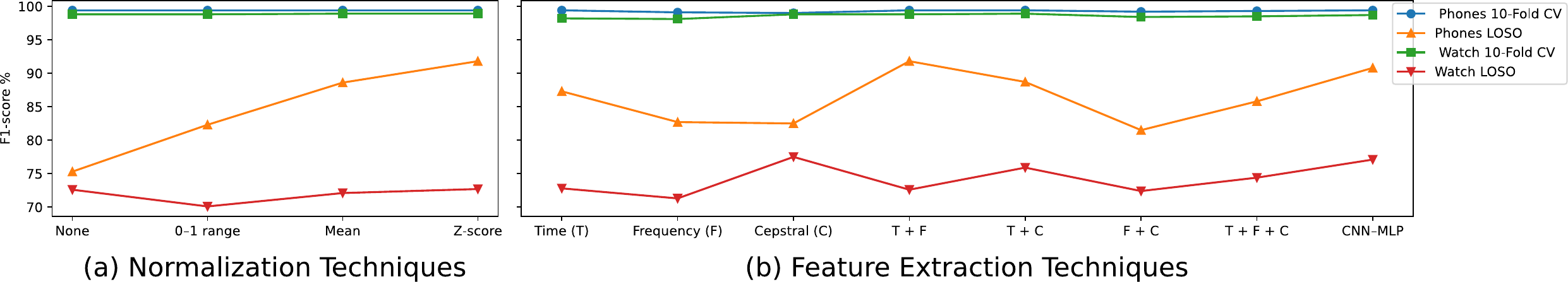"}
		\centering
		\caption{Reported F1-Score of a RF classifier from San-Segundo et al. \cite{san2018robust}. (a): different normalization techniques. (b): feature extraction techniques after z-score normalization.}
		\label{fig:segundo2018}	
	\end{figure}

	San-segundo et al. \cite{san2018robust} analysed different HAR approaches, such as discriminative, generative and deep learning algorithms, i.e. RF, Hidden Markov Models (HMM), and two hybrid models, CNN-MLP and CNN-LSTM. They used the Heterogeneity Human Activity Recognition (HHAR) Dataset \cite{stisen2015smart}, which contains accelerometer and gyroscope data captured using smartphones and smartwatches. The data were segmented using \textit{fixed-width sliding windows} of three seconds, with a two seconds overlap. The evaluation was using random stratified 10-fold CV and three different versions for leave-one-out (i.e., user, model, device) evaluation. Figure \ref{fig:segundo2018} depicts the F1-Score reported for each experiment. There is a significant drop in performance when the evaluation is done using the LOSO technique. The performance decrease is seen for both, smartphones (Figure \ref{fig:segundo2018} orange line, $\blacktriangle$ marker) and smartwatches (Figure \ref{fig:segundo2018} red line, $\blacktriangledown$ marker) data. This shows the bias introduced by sliding windows and random 10-fold CV. 
	%In order to show the influence of the bias introduced in the model because of the violation of the statistical independence assumption between samples, we focus only on the metrics reported for the 10-fold CV and LOSO evaluations. The F1-Score from the reported results of each experiment is shown in Figure \ref{fig:segundo2018}. As can been seen from the figure, there is a significant drop in performance when the evaluation is done using the LOSO technique. The performance decrease is seen for both, smartphones (Figure \ref{fig:segundo2018} orange line, $\blacktriangle$ marker) and smartwatches (Figure \ref{fig:segundo2018} red line, $\blacktriangledown$ marker) data.
	
	%%%% POSIBLY EXCLUDE
	Wang et al. \cite{wang2018eating} present an eating detection method based on data collected from a triaxial accelerometer attached to the temporalis muscle area. Besides eating, they collect data for reading, speaking, standing, sitting, walking, drinking and coughing activities. They propose a binary classification problem where eating is the positive class and any other activity represents the negative class. The data is segmented following a non-overlapping window approach of fixed length and used to extract different features from the time and frequency domain. The trained classifiers were: DT, NN, MLP, SVM and weighted SVM. The evaluation was performed using three CV techniques: 10-fold CV, one model per user, and LOSO CV. For the 10-fold CV, data from all users was joined and then split in 10 random folds. For the single-user evaluation data from each user represented a single dataset, then each user's dataset was split following the conventional 10-fold CV. Finally, for the LOSO CV, data from each user was used once as test set for every experiment. Not surprisingly, there is a considerable drop in performance when using LOSO CV (89\%) compared to the results reported for 10-fold (94.4\%) and Single-user CV (96.12\%). This confirms that even for non-overlapping windows, a random k-fold CV evaluation may produce overestimated results because of the temporal dependence between windows for long duration activities (e.g., eating) \cite{hammerla2015let, dehghani2019quantitative}. 
	
	%Figure \ref{fig:wang2018} shows the reported results for each evaluation technique. 
	
	% \begin{figure}[!bh]
		% 	\includegraphics[width=0.90\columnwidth]{"images/wang2018.pdf"}
		% 	\centering
		% 	\caption{Reported results from Wang et al., in \cite{wang2018eating}.}
		% 	\label{fig:wang2018}	
		% \end{figure}
	
\begin{figure}[!th]
		\includegraphics[width=0.75\columnwidth]{"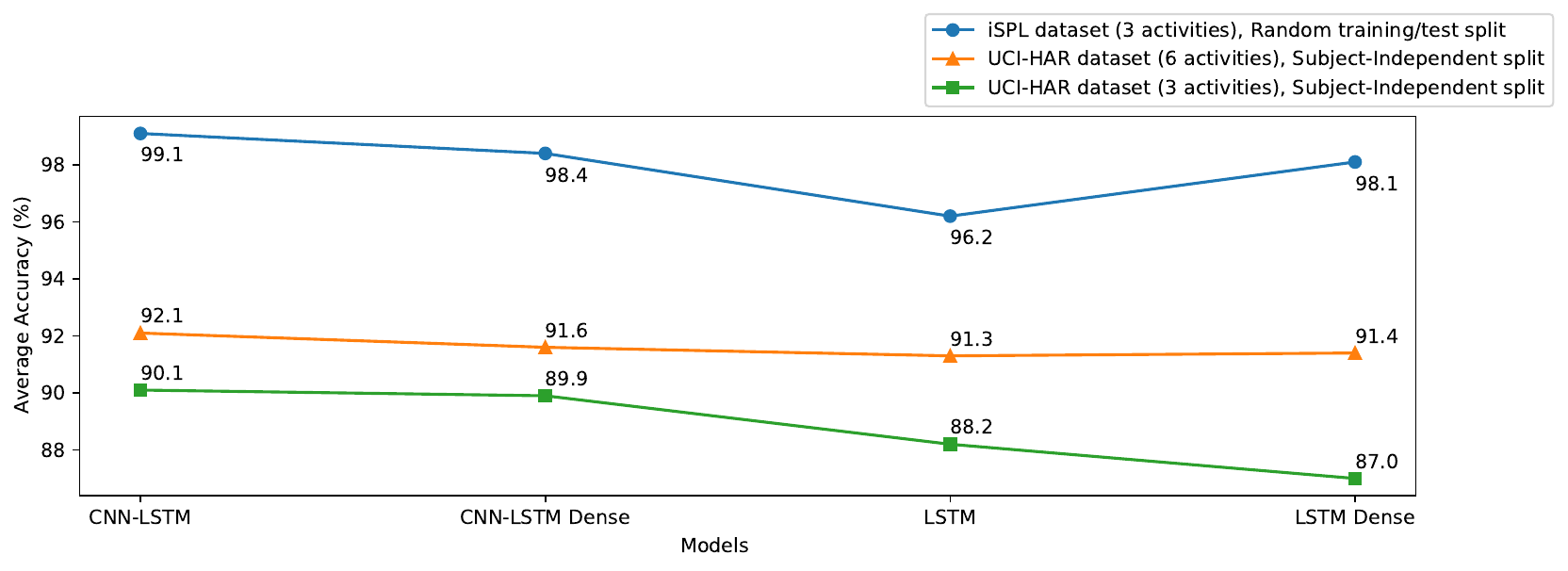"}
		\centering
		\caption{Reported results from Mutegeki and Han \cite{mutegeki2020cnn}.}
		\label{fig:mutegeki2020}	
	\end{figure}

	Mutegeki and Han \cite{mutegeki2020cnn} presented a CNN-LSTM approach for HAR. This work used the public UCI-HAR \cite{anguita2013public} dataset, and iSPL dataset, a private one created in their lab. The UCI-HAR dataset contains triaxial accelerometer and gyroscope data from a smartphone strapped to the waist of a subject. The experiments were carried out by 30 subjects performing six activities. The training and test sets were created following a subject-independent approach, where 70\% of the users were randomly chosen for training, and the 30\% different users were used for testing. The iSPL dataset contains triaxial raw accelerometer and gyroscope signals collected from a wearable sensor attached to the left-hand wrist of four subjectsperforming the Standing, Sitting, and Walking activities. In both cases, the data were segmented using fixed-width sliding windows of 2.56 sec with a 50\% overlap. For iSPL data, the windows were \textit{randomly split} in training and testing sets using a 80:20 ratio. The models evaluated were based on different combinations of CNNs and LSTMs. The results shown in this publication are the average accuracy of each experiment when classifying all the activities for each dataset. We added the results of the classification accuracy of UCI-HAR only for the common activities between both datasets (Figure \ref{fig:mutegeki2020} green line, square markers $\blacksquare$). The classification accuracy on the iSPL dataset is notably better for all the experiments, including all activities and only common activities. Therefore, it is likely that the \textit{random training/test split} used with the iSPL dataset, did not follow a subject-independent approach (contrary to UCI-HAR), and thereby the results show an overestimation of the models' accuracy. 
	
	Ni et al. \cite{ni2020leveraging} propose Stacked Denoising Autoencoders for HAR and SMOTE resampling for dealing with class imbalance. They used their own dataset collected from ten adults using wearable accelerometer and gyroscope devices. They also used the UCI-HAR, the chest-mounted accelerometer data \cite{casale2011human}, and the UCI Daily and Sports Activities datasets for evaluation. The data segmentation was performed using a 512 samples with a 50\% overlap. After resampling and data standardization, ``the whole dataset was divided into training set, validation set, and testing set with the ratio of 6:2:2 randomly'' \cite{ni2020leveraging}. Given this approach, it is expected that the reported accuracies are overestimated. Also, the methodology description implies that data standardization was performed before training/test splits. But it is crucial that normalization parameters (min, max, mean, variance, etc.) are learned only from the training data, and then used to normalize the training and test sets \cite{gholamiangonabadi2020deep}, avoiding training/test set data contamination.
	
	Gupta \cite{gupta2021deep} proposed a deep learning approach for HAR using the WISDM benchmark dataset \cite{weiss2019smartphone}. This dataset contains accelerometer and gyroscope data collected from 51 subjects using smartphones and smartwatches while performing 18 activities. The dataset is organized in flat files, 4 files per user. Gupta merged the acceleration and gyroscope data from all users into a single dataset for smartphones, and a single one for smartwatches. Both datasets were segmented using a \textit{sliding window of 10 seconds and 50\% overlap} and split in training, validation and test sets in a 60:20:20 ratio. In all the experiments\footnote{\scriptsize Gupta (2021) source code: \href{https://github.com/guptasaurabh777/HAR}{https://github.com/guptasaurabh777/HAR}}, the splits were performed using the train\_test\_split function from the scikit-learn python library with the boolean parameter `shuffle' set to `True', which causes the data is shuffled before splitting. The reported results are an accuracy of 96.54\% for the smartwatch data and 90.44\% for the smartphone dataset. These results are based, once again, on overlapping sliding windows and random training/test sets split, which implies a performance overestimation. 
	
\begin{figure}[!th]
		\includegraphics[width=0.75\columnwidth]{"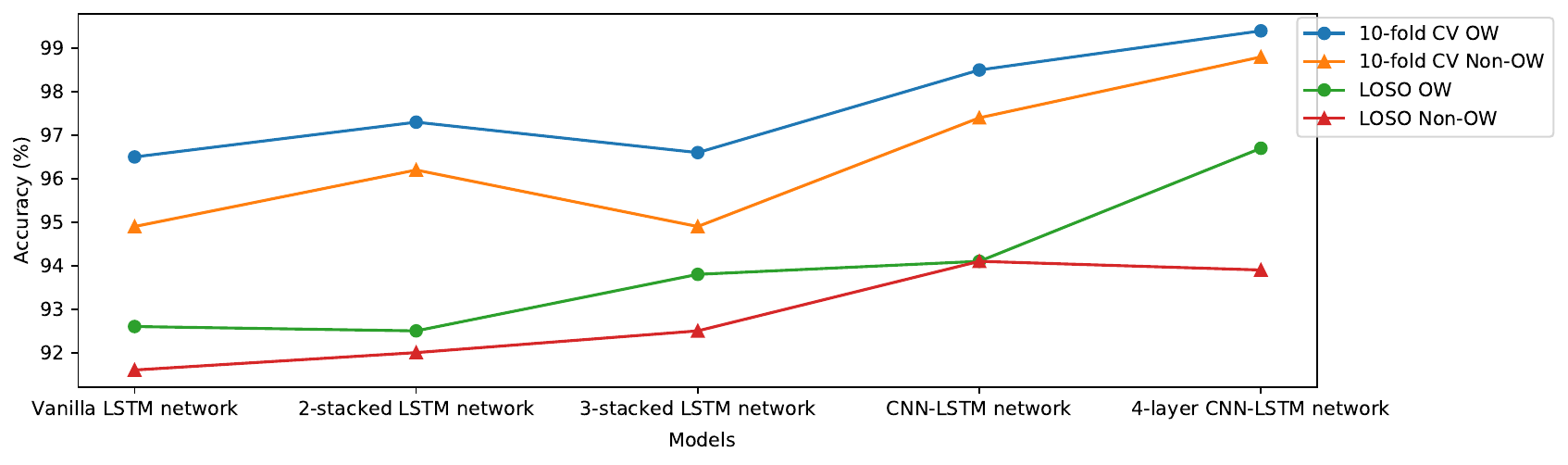"}
		\centering
		\caption{Reported results from Mekruksavanich and Jitpattanakul \cite{mekruksavanich2021lstm}.}
		\label{fig:mekruksavanich2021}	
	\end{figure}

	Mekruksavanich and Jitpattanakul \cite{mekruksavanich2021lstm} present five different LSTM Networks configurations for HAR using the UCI-HAR dataset. They applied the overlapping and non-overlapping windows approaches for data segmentation. However, they did not evaluate the models using the default training/test sets provided, which are user-independent. In this study, the models were evaluated using the conventional 10-fold CV and the LOSO CV techniques. The results of the experiments show the performance overestimation when using a 10-fold CV vs. following a LOSO evaluation approach. Most importantly, this work evidences that the overestimation occurs using either overlapping and non-overlapping windows for data segmentation (See Figure \ref{fig:mekruksavanich2021}). The overestimation is also confirmed comparing the results obtained with LOSO against those reported in \cite{mutegeki2020cnn} for UCI-HAR dataset (see Figure \ref{fig:mutegeki2020}).
	
	Li et al. \cite{li2021tribogait} presented a Deep Learning approach for HAR from electrical signals produced by a triboelectric gait sensor. Two gait sensor modules were attached to a carpet to produce the signals collected while 8 subjects performed 5 different activities. The data were segmented using a \textit{3 seconds sliding window with a step size of 0.02 seconds}. According to the authors, such overlap is ``to preserve temporal information and provide more redundancy''. Then, the window data samples were \textit{randomly split} in a 80:20 ratio. Once again, the bias introduced with the followed approach is overlooked. 
	
	Bouchabou et al. \cite{bouchabou2021fully} presented a Fully Convolutional Network combined with Natural Language Processing (NLP) techniques for HAR. They used the Aruba and Milan datasets from the CASAS benchmark dataset \cite{cook2012casas}. These datasets are of different nature than those already described. They include mostly binary sensors (e.g., motion, contact) mounted in a smart home testbed environment. In this work, each sensor event is treated as a word and activity sequences as sentences in text and vector representations of those words are learned using NLP techniques. The data segmentation follows an overlapping sliding windows approach with different windows sizes with a step size of 1 sensor reading for all cases. With this approach, there is more than 95\% overlap. Finally, the whole dataset was partitioned in a random stratified 70:30 ratio. They claim a ``better generalization'' obtained by means of a random shuffle before splitting. The F1-scores reported for the Aruba dataset are 99\% for a 25-window and 100\% for all other windows sizes. Likewise, for the Milan dataset, the F1-scores are above 94\% for all windows sizes.
	
	G{\'o}mez Ramos et al. \cite{ramos2021daily} presented another work on HAR using the CASAS data \cite{cook2012casas}. A sliding window of 60 records was used for data segmentation and randomly split in a 70:10:20 ratio for training, validation and testing respectively. Windows were stacked in pairs before randomizing the splits, according to the authors, to keep the temporal sequence of events. However, the pairs of windows were still randomly shuffled, thus allowing for consecutive pairs to appear in both training and test sets. This work reported an accuracy and f-score above 95\%. Once again, the approach followed in this work shows a model performance overestimation, comparable to the previously reported in \cite{bouchabou2021fully}.
	
	Zimbelman and Keefe \cite{zimbelman2021development} presented a RF approach for HAR to recognize activities performed by rigging crew workers on cable logging operations. The data were collected using a smartwatch that recorded accelerometer data at 25 Hz, while the workers set and disconnect log chokers at five logging sites. The experiments included fixed-size sliding windows that go from 1 to 15 seconds with overlapping up to 90\%. The whole dataset was split following a random stratified approach with a 67:33 ratio. The best reported accuracies $\approx 89\%~$ were those obtained from sliding windows with 90\% overlap. This shows that the higher overlap, the higher overestimation.
	
	Yan et al. \cite{yan2022deep} propose a HAR model based on GNNs. They used data from the MHEALTH, PAMAP2 and their own dataset TNDA-HAR. The raw input data is transformed into a graph representation based on the Pearson correlation coefficient between the sensors channels signals. Each channel represents a graph vertex, and a correlation of a pair of vertices above 0.2 implies an edge. A GNN model is trained to encode the graph, followed by two fully connected layers as the final classifier. The authors report accuracies of 98.18\% for PAMAP2, and 99.07\% for MHEALTH. But those accuracies are overestimated due to the use of random training/test sets split, which is seen in the source code provided by the authors.
	
	Wang et al. \cite{wang2022sensor} compared the effect on performance of different data augmentation methods using UCI-HAR, USC-HAD \cite{malekzadeh2018protecting}, MotionSense \cite{zhang2012usc} and MobiAct \cite{chatzaki2017human} datasets. They segmented the data using sliding windows with 12.5\% to 50\% overlap. In some experiments they randomly split the data into training/test sets, but they also followed a subject-independent 5-fold CV for the MotionSense dataset. The F1-scores reported for all the experiments using randomly partitioned data were: UCI-HAR 98.28, MotionSense 99.35, USC-HAD 92.28 and MobiAct 98.32. In accordance to previous findings, in the experiments with MotionSense dataset but with a subject-independent CV, the F1-score dropped from 99.35 to 92.10.
	
	Huang et al. \cite{huang2022channel} proposed Channel-Equalization-HAR, a variation of the normal CNNs. They used UCI-HAR, OPPORTUNITY \cite{roggen2010collecting}, UniMiB-SHAR, WISDM, PAMAP2, and USC-HAD datasets. The data was segmented using sliding windows of different sizes with 30\%, 50\%, 78\% overlap. Although some of those datasets already provide training, validation and test sets following a subject-independent approach, the authors mention that they split all datasets in a 7:1:2 ratio. This suggests that the data was merged and randomly split following the specified ratio later on. The reported F1-score for UCI-HAR was 97.12\%, similar to the ones reported in \cite{mekruksavanich2021lstm, wang2022sensor} which followed a random partition of train/test sets. Likewise, the reported accuracy on the WISD dataset was 99.04\%, even higher than the one reported in \cite{gupta2021deep} which followed the biased approach. The same can be observed for the USC-HAD dataset with a reported F1-score of 98.93\%, higher than reported in \cite{wang2022sensor} which also has the problem. With the PAMAP2 dataset, the reported F1-score is 92.18\% which is similar to our own experiments using random training/test set splits, shown later in Section \ref{sec:experiments}.
	
	Luo et al. \cite{luo2021binarized} proposed a Binarized Neural Network for HAR, which allows to move the computation to the edge. They used the Radar HAR dataset \cite{luo2019kitchen}, UCI-HAR and UniMib-SHAR datasets. They applied sliding windows data segmentation followed by a random training/test split, in a 80:20 ratio for the Radar HAR dataset, and a 70:30 ratio for the UCI-HAR and UniMib-SHAR datasets. The F1-score reported for the Radar HAR dataset is 98.6\%. The reported F1-score for the UCI-HAR dataset is 98.1\%, similar to the ones reported in \cite{mekruksavanich2021lstm, wang2022sensor, huang2022channel} which followed the same biased approach. The reported F1-score for the UniMib-SHAR dataset is 93.3\%, even higher than the one reported in \cite{micucci2017unimib} which is also overestimated.
	
	Wu et al. \cite{wu2023novel} proposed a spatio-temporal LSTM model using data from a pedal wearable device attached to the footwear's tongue area. The approach combines a GNN model for the spatial features and a LSTM for the temporal patterns to recognize five different activities. A sliding window of 200 samples was used for data segmentation. The segmented data was randomly split into training/test sets. The reported results show a perfect F1-score of 1.0 for the Sitting and Down the Stairs activities, 0.96 and 0.97 for Standing and Walking respectively, and 0.83 for Up the Stairs. Once again, the followed approach shows overestimated results.
	
	Garcia-Gonzalez et al. \cite{garcia2023new} used their own dataset containing accelerometer, gyroscope, magnetometer and GPS data from smartphones. They evaluated different traditional ML algorithms: SVM, DT, MLP, NB, k-NN, RF and Extreme Gradient Boosting (XGB). Data segmentation and feature extraction were performed using sliding windows from 20 to 90 seconds with 1 second step size, i.e, at least 95\% overlap. Then, a stratified k-fold CV is used to evaluate the different models. This method tries to keep a similar distribution of the samples per class within each fold, but it does not prevent that consecutive windows are assigned to two different folds. Hence, the reported accuracy, 92\%, is overestimated.
	
	\section{Unbiased model evaluation}
	\label{sec:experiments}
	
	This section describes our experiments to compare biased vs. unbiased evaluation methods. Here, the focus is to show how the same approach can lead to a considerable difference in accuracy depending on the followed data segmentation and training/test sets partition strategies.
	
	\subsection{Datasets}
	
	We used different types of datasets to evaluate whether the problem described in section \ref{sec:problem} affects in the same way datasets with different data modality. We used the \textbf{MILAN} dataset \cite{cook2009assessing} from the CASAS \cite{cook2012casas} benchmark dataset collection\footnote{\text{CASAS benchmark dataset collection: } \href{http://casas.wsu.edu/datasets/}{http://casas.wsu.edu/datasets/}}, which contains binary data from motion and contact sensors, and the \textbf{PAMAP2} \cite{reiss2012intro} and \textbf{MHEALTH} \cite{banos2014mhealthdroid} datasets, which contain accelerometer, gyroscope and magnetometer data from wearable on-body sensors. 
	
	\paragraph{\textbf{MILAN}} The data were collected in a smart home testbed environment. The sensors mounted in the house included 28 motion sensors, 3 door contact sensors and 2 temperature sensors. In this dataset the beginning and the end of the activities are annotated. It is assumed that all the samples enclosed by the \textit{begin} and \textit{end} markers belong to the annotated activity. Conversely, those samples which are outside the markers has no label associated to them. Hence, such data samples have been annotated as the \textit{``other''} class. This dataset is highly imbalanced, the \textit{``other''} being the dominant class.
	%Figure \ref{fig:milandata} shows an overview of the Milan dataset composition.
	
	% \begin{figure}[!bh]
		% 	\centering
		% 	\begin{subfigure}{0.18\textwidth}
			% 		\centering
			% 		\includegraphics[width=0.8\textwidth]{"images/data_milan.pdf"}
			% 		\caption{Raw data}
			% 		\label{fig:milanraw}
			% 	\end{subfigure}%
		% 	\begin{subfigure}{0.32\textwidth}
			% 		\centering
			% 		\includegraphics[width=0.8\textwidth]{"images/activities_dist.pdf"}
			% 		\caption{Samples distribution}
			% 		\label{fig:milandist}
			% 	\end{subfigure}
		% 	\caption{Overview of the Milan dataset.}
		% 	\label{fig:milandata}
		% \end{figure}

	\paragraph{\textbf{PAMAP2}} This dataset contains data collected from 9 subjects using IMUs attached to the wrist, chest and ankle while performing everyday, household and sport activities. Each sensor includes two accelerometers, one gyroscope and one magnetometer producing 3-axial data at a sampling rate of 100Hz. This dataset includes twelve main activities and six optional. In our experiments we only used the main twelve activities: \textit{Lying down, Sitting, Standing, Walking, Running, Cycling, Nordic Walk, Walking Upstairs, Walking Downstairs, Vacuum Cleaning, Ironing} and \textit{Rope Jumping}.
	
	\paragraph{\textbf{MHEALTH}} This dataset contains data of 10 volunteers performing twelve physical activities: \textit{Standing, Sitting, Lying down, Walking, Climbing stairs, Waist bends forward, Arms up, Knees Bending, Cycling, Jogging, Running} and \textit{Jumping}. The sensors were placed at subjects' chest, right wrist and left ankle. The data comprise 3-axis accelerometer, gyroscope and magnetometer signals collected at a sampling rate of 50Hz. The sensor placed at the chest also provides 2-lead ECG measurements, but those data points were not used in the experiments. 
	%Figure \ref{fig:imu-dist} shows an overview of the PAMAP2 and MHEALTH datasets composition.
	
	% \begin{figure}[!th]
		% 	\centering
		% 	\begin{subfigure}{0.25\textwidth}
			% 		\centering
			% 		\includegraphics[width=1\textwidth]{"images/pamap2_dist.pdf"}
			% 		\caption{PAMAP2}
			% 		\label{fig:pamap-dist}
			% 	\end{subfigure}%
		% 	\begin{subfigure}{0.25\textwidth}
			% 		\centering
			% 		\includegraphics[width=1\textwidth]{"images/mhealth_dist.pdf"}
			% 		\caption{MHEALTH}
			% 		\label{fig:mhealth-dist}
			% 	\end{subfigure}
		% 	\caption{Samples distribution of PAMAP2 and MHEALTH Datasets.}
		% 	\label{fig:imu-dist}
		% \end{figure}	
	
	\subsection{Data segmentation and feature extraction}
	
	We perform data segmentation using a fixed-size sliding windows approach as proposed in \cite{krishnan2014activity} and further analyzed in \cite{quigley2018comparative}. 
	
	\paragraph{\textbf{MILAN}} This dataset mainly comprises binary motion and contact sensors. We used a window of $k$ sensor events because these type of sensors do not fire a sample at a constant rate.	The window size was 30 with a step-size of 1, based on empirical evaluation as presented in \cite{krishnan2014activity, wang2021multi}. The windows were created based on collection date, this approach facilitates the independence between training and test sets during model evaluation. The last sensor event in the window defines the label and the preceding events in the window define its context. Following this approach we can have a prediction every time a new sensor event arrives, achieving a near-real time recognition. On the other hand, there might be multiple transitions within a single window, or the last sensor event may be dominated by sensors from previous events. We alleviate these problems using a weighting mechanism based on the Mutual Information (MI) probability as described in \cite{krishnan2014activity}. MI measures the chance of two sensors being activated consecutively. It is likely that if two motion sensors are located in the same room, the triggering of one will follow the activation of the other. The same applies between contact and motion sensors. Hence, sensors that are close to each other will have a higher MI value than those that are far apart.
	
	Then, feature vectors are calculated for each window following the approach presented in \cite{krishnan2014activity, wang2021multi}. We transformed the \textit{hour-of-day} and \textit{day-of-week} to sine and cosine pairs to capture the equidistant relation between time-based cyclical values \cite{ramos2021daily}. The activations, in the case of motion sensors, are the number of times each sensor in the window is ON. For door contact sensors, it is the number of times the sensor appears in the window sequence. Instead of using the counts of activations for each sensor directly, this value is weighted by the MI probability, refer to \cite{krishnan2014activity} for more detailed information on MI calculation. 
	
	Other features considered are the sensor's status, defined from {0,1} for motion and contact sensors, the average temperature in the window, the window entropy, and the the sparsity of the window, calculated as \textit{sparcity = 1 - probability\_non\_zero\_values}. In addition, based on the floor plan of the apartment, the location of the last sensor event in the window and the location	of the last motion sensor triggered are included in the feature set.
	
	\paragraph{\textbf{PAMAP2}} This dataset was segmented using sliding windows of 5.12 seconds with 1 second shift, following the approach of its original publication \cite{reiss2012intro}. Since the data were sampled at 100Hz, the windows span 512 sample readings with step-size of 100 samples. This data was used to train a classification model based on GNNs. Therefore, the training data was transformed to a graph representation where the vertices correspond to the different channels of the sensors' signals. The edges are defined by means of the Pearson's Correlation Coefficient between the channels, where a correlation threshold above 0.2 implies an edge between two channels.
	
	\paragraph{\textbf{MHEALTH}} Following the protocol in \cite{anguita2013energy, anguita2013public}, we segmented the data using a sliding window of 2.56 seconds with 50\% overlap. Since this dataset was sampled at 50Hz, the windows have 128 samples with 64 samples overlap. For feature creation, we first transformed the window data into a graph representation in the same way as described for PAMAP2 dataset. Then, the activity graphs were used to train a GNN-based classification model.
	
	\subsection{Classification models and evaluation strategies}
	
	We trained two different types of models to evaluate the effect of sliding window segmentation and random training/test splits when using classification models of different nature. We trained a RF classifier to categorize human activities based on binary sensor data, and we used a GNN-based classification model for data from on-body IMU sensors.
	
	\paragraph{\textbf{MILAN}} After feature extraction, the data were partitioned into training and test sets in a 80:20 ratio. The model was trained and finetuned with the conventional 5-fold CV approach over the training data. Then, with the best parameters obtained from CV, the model was retrained on the whole training set and evaluated on the hold-out 20\%. First, we partitioned the data completely random using the \textit{train\_test\_split} function, from the scikit-learn python library \cite{scikit-learn}, with the \textit{shuffle} parameter set to \textit{True}. Then, in a second experiment, we used the \textit{StratifiedGroupKFold} CV scheme from the scikit-learn python library. This scheme splits the data into folds with non-overlapping groups, where percentage of samples for each class is preserved. In our case, the groups were determined by the \textit{collection\_date} of the raw data samples. 
	
	\paragraph{\textbf{PAMAP2 and MHEALTH}} Most of the studies presented in section \ref{sec:evidence} followed a conventional Deep Learning approach for HAR, e.g., CNN, LSTM or a combination thereof. Following the approach of Yan et al. \cite{yan2022deep}, we wanted to confirm whether the bias introduced by the problem described in section \ref{sec:problem} also affects a different type of Deep Learning models. Therefore, we trained a 3-layer GNN implemented using the Graph Convolution presented in \cite{Morris:2019weisfeiler}, followed by a two fully connected layers and softmax layer for final classification. We split our segmented data in a 6:2:2 ratio used for training, validation, and final model evaluation, respectively. Similarly as we did for Milan dataset, we first partitioned the data randomly, and later on using the StratifiedGroupKFold approach, determined by \textit{subject\_id}. We did the same for both, PAMAP2 and MHEALTH datasets. In the case of PAMAP2, subjects 101 and 107 were used for validation, subjects 103 and 105 for testing, and the remaining subjects for training. For MHEALTH dataset, subjects 6 and 10 were used as validation set, subjects 2 and 9 for testing and the rest for training. After hyper-parameter optimization and finding the best parameters combination, the model was updated with the entire training data, including the training and validation subsets. Finally, the model was evaluated using the hold-out test set.

	\subsection{Results and discussion}
	\label{sec:discussion}
	
	Similarly to previous studies presented in Section \ref{sec:evidence}, the performance obtained with a random training/test split significantly outperformed the one obtained with the grouped data. Table \ref{tab:comp-results}  compares the classification performance of the three datasets on randomly partitioned data vs. group-based splits. These results clearly show that the performance of the models on randomly partitioned data is highly overestimated, independently of the dataset and classification model used. It is important to point out that the same model was trained on both, randomly partitioned and group-based partitioned data. This shows that the model performance is clearly affected by the data partition strategy that follows the window-based data segmentation. Therefore, evidencing that window-based data segmentation and random data splits leads to misleading results, as described in Section \ref{sec:problem}.
	
	\begin{table}[thb]
		\centering
		\caption{Classification performance comparison on MILAN, PAMAP2 and MHEALTH datasets.}
		\label{tab:comp-results}
		\resizebox{\columnwidth}{!}{%	
			\begin{tabular}{lcccccc}
				\toprule
				\multirow{2}{*}{Partitioning} & \multicolumn{2}{c}{ \footnotesize MILAN \scriptsize(RF)} & \multicolumn{2}{c}{ \footnotesize PAMAP2 \scriptsize (GNN)} & \multicolumn{2}{c}{ \footnotesize MHEALTH \scriptsize (GNN)} \\
				\cmidrule(lr){2-3} \cmidrule(lr){4-5} \cmidrule(lr){6-7} 
				& \scriptsize b. acc  &  f1-score & \scriptsize b. acc & \scriptsize f1-score & \scriptsize b. acc & \scriptsize f1-score \\
				\midrule
				Random &  92.91 & 95.17 & 90.90  & 91.97  & 97.61 & 97.95  \\
				Group-based & 60.06  & 56.64 & 85.50 & 85.73 & 85.46 & 85.77 \\
				\bottomrule
			\end{tabular}%
		}
	\end{table}

\begin{figure}[!th]
		\includegraphics[width=0.75\columnwidth]{"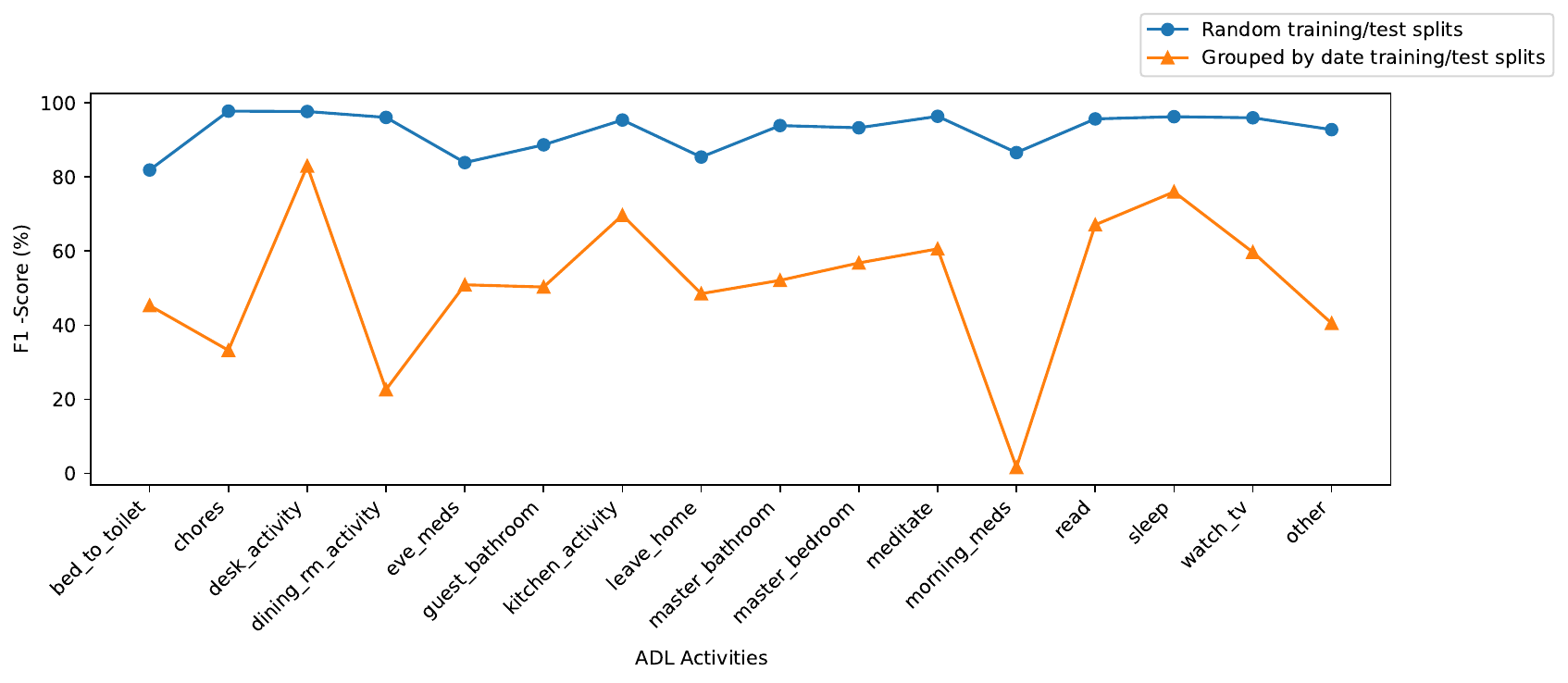"}
		\centering
		\caption{F1-Score obtained for each activity with the random and group-based data partitioning schemes on the Milan dataset.}
		\label{fig:experiments}	
	\end{figure}

	\paragraph{\textbf{MILAN}} The results obtained with the Milan dataset from the randomly split experiment were a balanced-accuracy of 92.91\% and a f1-weighted score of 95.17\%. Those values dropped to a 60.06\% balanced-accuracy and 56.64\% f1-weighted score following the group-based partitioning scheme. Figure \ref{fig:experiments} shows that the performance with random splits is higher in all sixteen activities. These results conform to those reported by Bouchabou et al. \cite{bouchabou2021fully} and by G{\'o}mez Ramos et al. \cite{ramos2021daily}.
	
\begin{figure}[H]
		\centering
		\begin{subfigure}{0.5\columnwidth}
			\centering
			\includegraphics[width=0.90\columnwidth]{"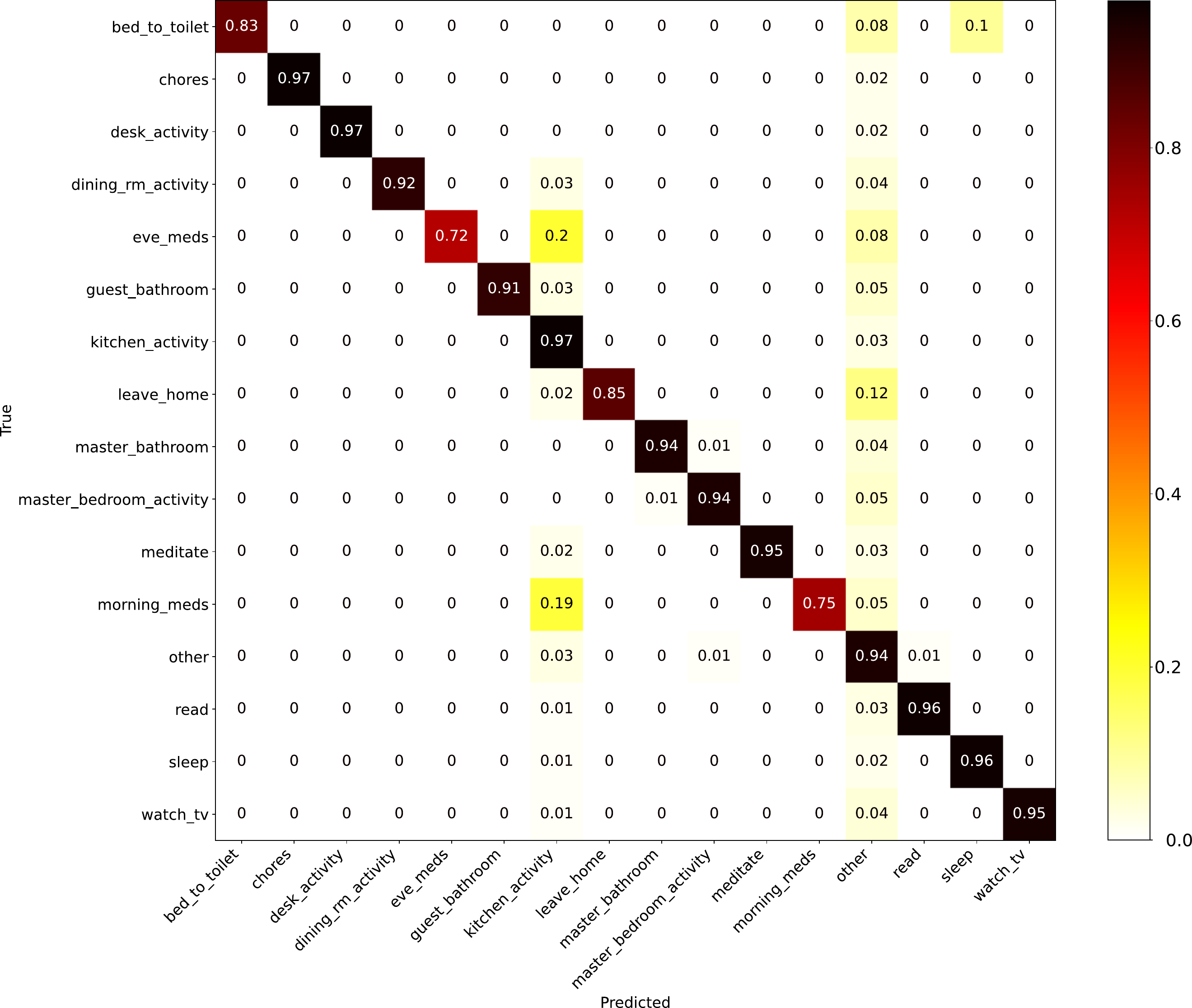"}
			\caption{Random training/test split}
			\label{fig:randomcv}
		\end{subfigure}%
		\begin{subfigure}{0.5\columnwidth}
			\centering
			\includegraphics[width=0.90\columnwidth]{"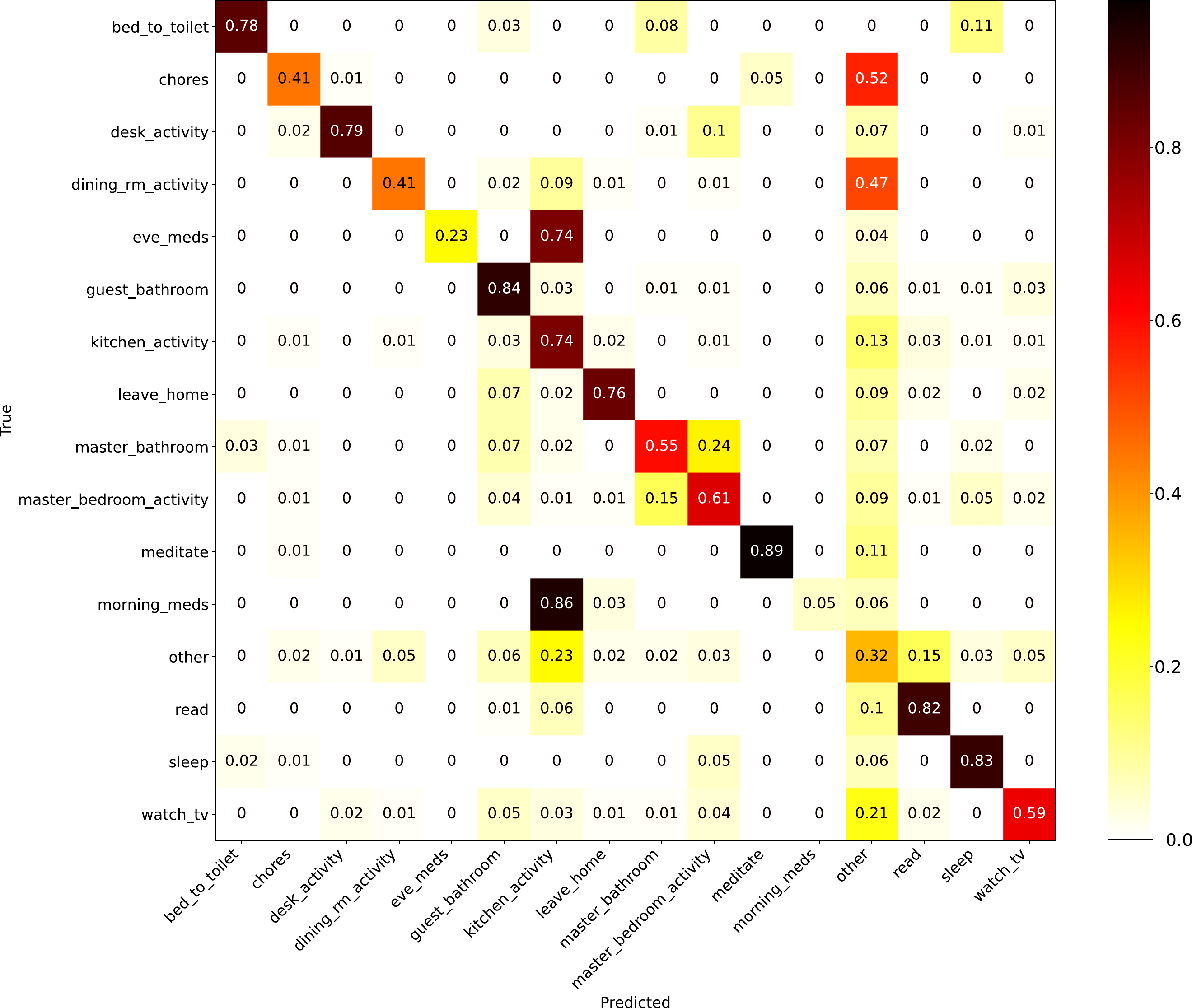"}
			\caption{Group partitioning}
			\label{fig:groupedcv}
		\end{subfigure}
		\caption{Confusion matrices of the RF classification model on the Milan dataset.}
		\label{fig:cmtrx}
	\end{figure}

	The confusion matrices obtained using both partitioning schemes are shown in Figure \ref{fig:cmtrx}. The results show that in both experiments the minority classes  are misclassified, i.e. \textit{morning\_meds}, \textit{evening\_meds}. The reason is because those activities occur in the same room; thereby, they trigger the same set of sensors. However, it is important to point out that the biased model classified the \textit{morning\_meds} and \textit{evening\_meds} with an accuracy over 72\%. On the contrary, the accuracy on the same classes for the group-based partitioned data is under 23\%. The same happens with the \textit{master\_bathroom} and \textit{master\_bedroom\_activity} classes, both occur in the main bedroom in the house. However, while the biased model correctly classifies each activity with an accuracy of 94\%, the corrected model classifies the same activities with an accuracy under 61\%.
	
	The data imbalance negatively affects the performance of the classifier but this effect is more acute in the corrected model. The \textit{``other''} is the majority class corresponding to the 32.8\% of the data. While the biased model produced an accuracy of 94\% on the \textit{``other''} class, the corrected model just obtains a 32\% accuracy on this pseudo activity. 
	
\begin{figure}[thb]
		\centering
		\begin{subfigure}{0.25\textwidth}
			\centering
			\includegraphics[width=0.75\textwidth]{"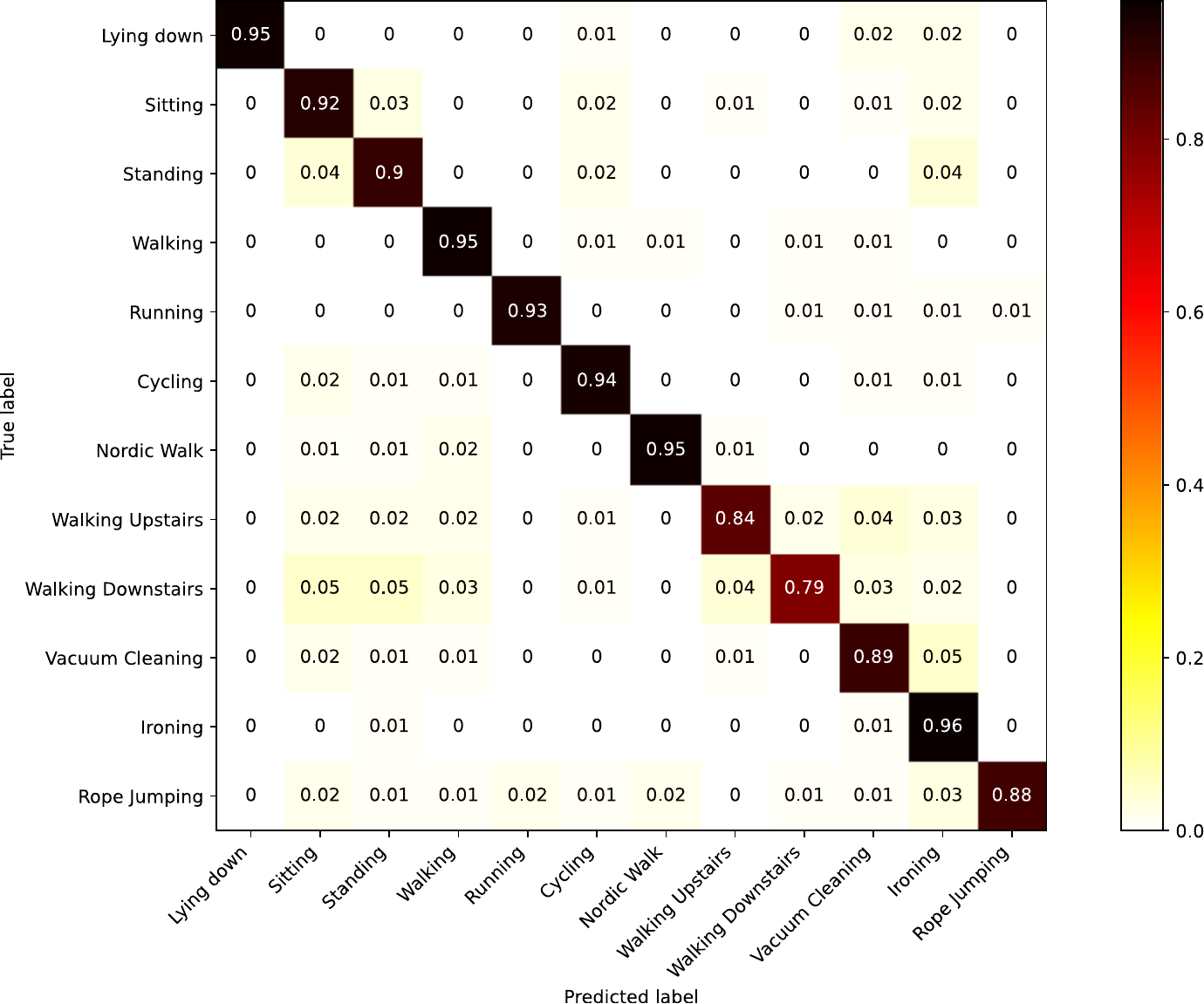"}
			\caption{Random training/test split}
			\label{fig:random-pamap2}
		\end{subfigure}%
		\begin{subfigure}{0.25\textwidth}
			\centering
			\includegraphics[width=0.75\textwidth]{"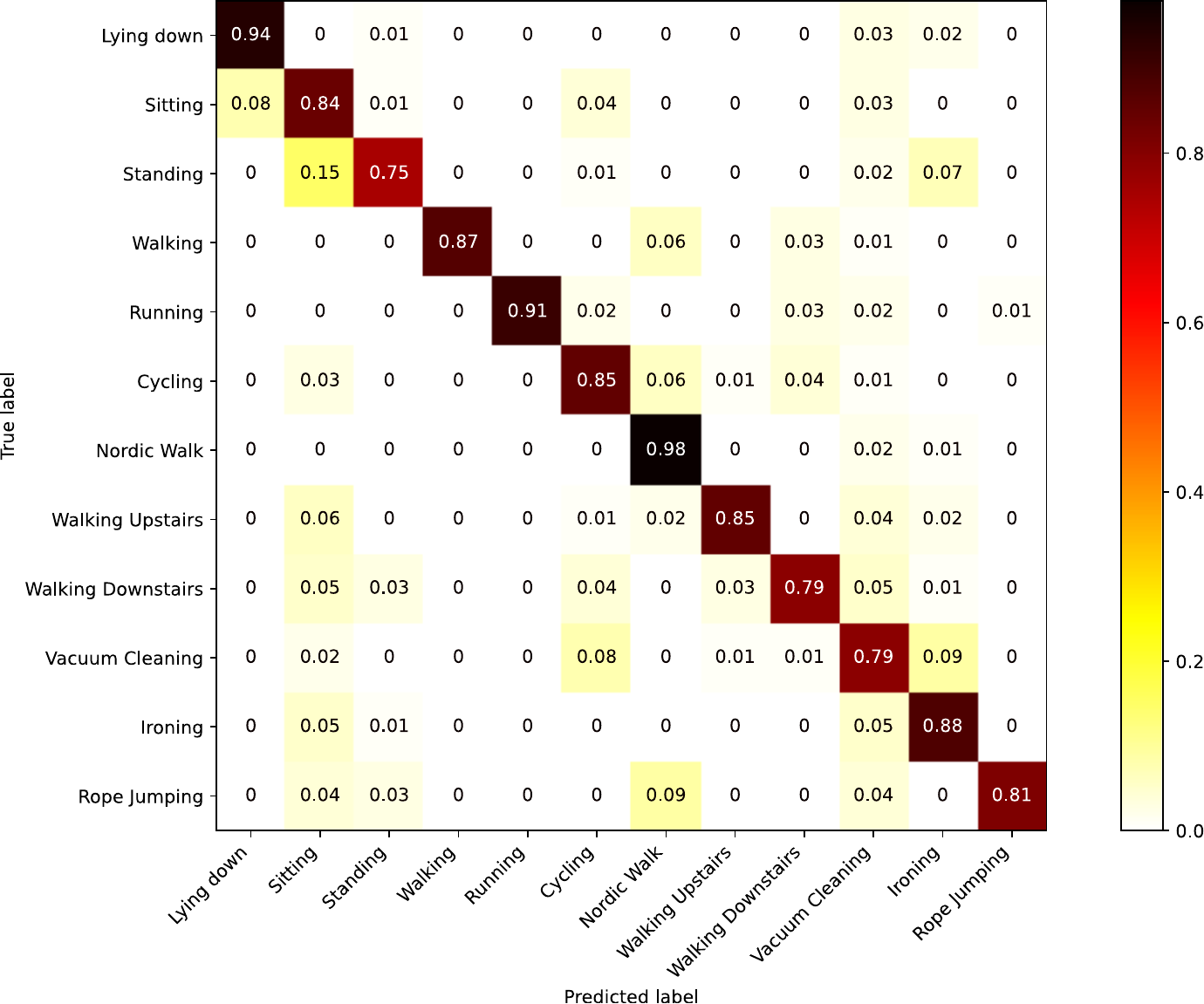"}
			\caption{Group partitioning}
			\label{fig:grouped-pamap2}
		\end{subfigure}
		\caption{Confusion matrices of the GNN classification model on the PAMAP2 dataset.}
		\label{fig:cmtrx-pamap2}
	\end{figure}

\begin{figure}[thb]
		\centering
		\begin{subfigure}{0.25\textwidth}
			\centering
			\includegraphics[width=0.75\textwidth]{"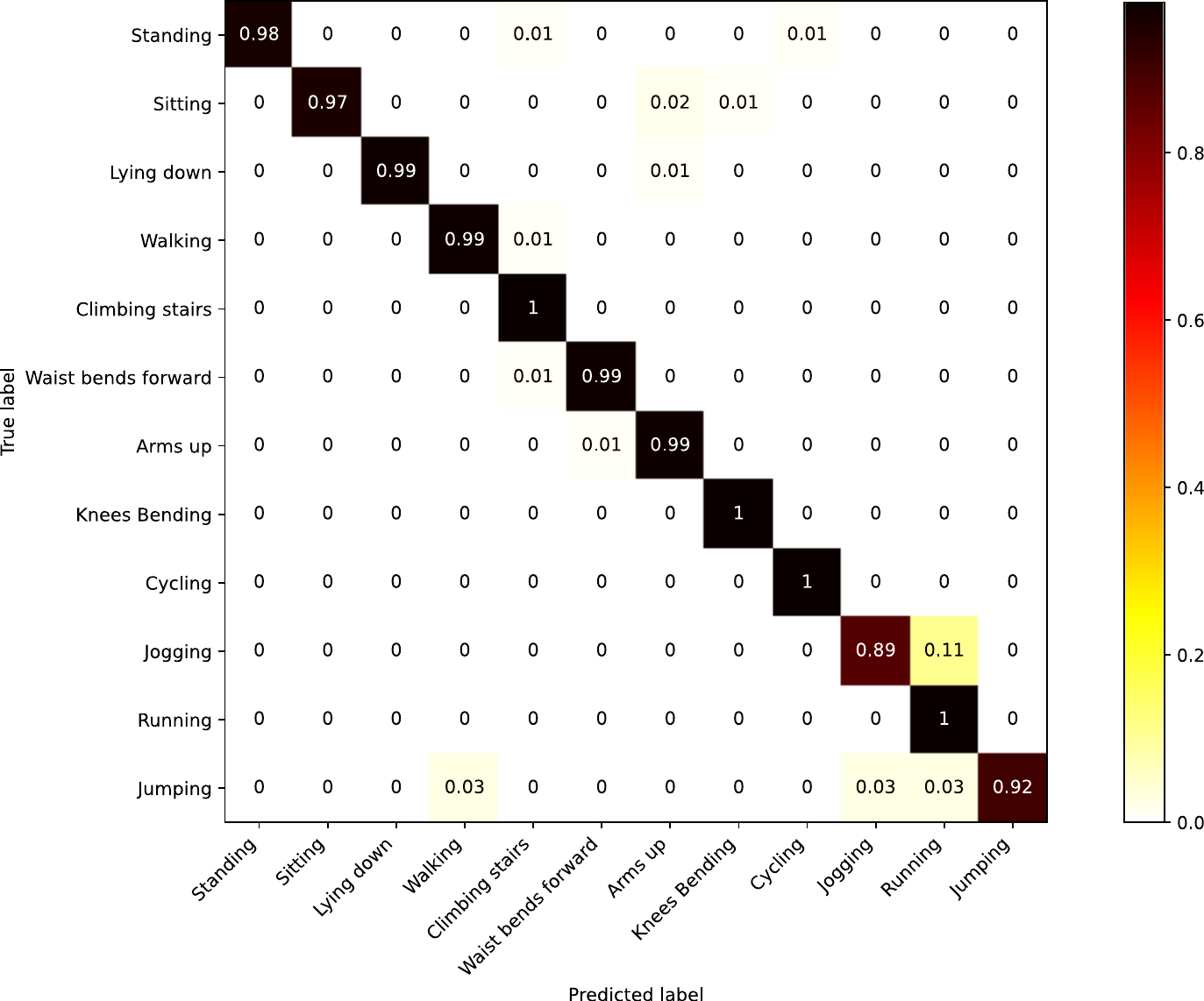"}
			\caption{Random training/test split}
			\label{fig:random-mhealth}
		\end{subfigure}%
		\begin{subfigure}{0.25\textwidth}
			\centering
			\includegraphics[width=0.75\textwidth]{"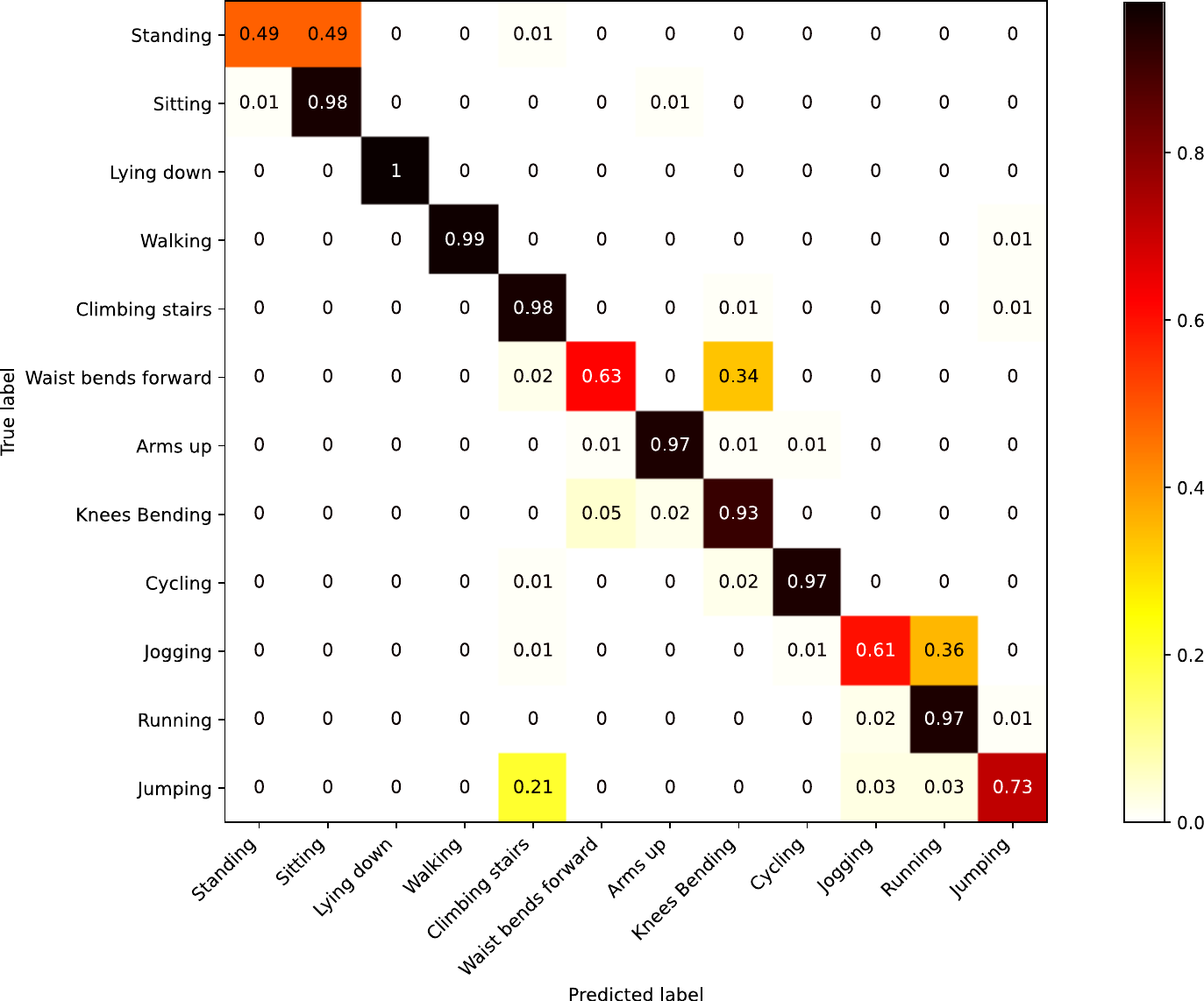"}
			\caption{Group partitioning}
			\label{fig:grouped-mhealth}
		\end{subfigure}
		\caption{Confusion matrices of the GNN classification model on the MHEALTH dataset.}
		\label{fig:cmtrx-mhealth}
	\end{figure}

	\paragraph{\textbf{PAMAP2 and MHEALTH}} Likewise with Milan dataset, the results obtained with the GNN-based classification model on PAMAP2 and MHEALTH datasets also show a performance overestimation when the model is trained and evaluated on randomly partitioned data. In the PAMAP2 dataset the accuracy and f1-score dropped from $\approx 92\%$ to $\approx 86\%$, when the data is partitioned following a group-based approach. In MHEALTH the performance decreased from $\approx 98\%$ to $\approx 86\%$  (See Table \ref{tab:comp-results}). This can also be observed from the confusion matrices of the GNN-based classification model on the PAMAP2 and MHEALTH datasets (Figure \ref{fig:cmtrx-pamap2} and Figure \ref{fig:cmtrx-mhealth}). These results confirm that the ``perfect'' accuracy reported by Yan et al. \cite{yan2022deep} is overestimated due to the windowing mechanism and random training/test set splits.	
	
	On the other hand, the results show that GNN-based classification models produce consistent results on both datasets. This can be observed from the confusion matrices where \textit{sitting} and \textit{standing} classes are misclassified in both, PAMAP2 and MHEALTH datasets. In the case of MHEALTH the model also confuses \textit{jogging} with \textit{running} activities, which is acceptable since the only difference between both is the intensity at which the activity is executed, and this mainly depends on how the user performs each activity.

	\section{Conclusions}
	\label{sec:conclusions}
	
	Our work systematically reviews HAR works with approaches that lead to model performance overestimation to raise awareness of the HAR community of this ongoing problem. Due to publications with biased overestimated results, fair approaches, with correct unbiased methods, may be disregarded due to the erroneously perceived low accuracy.   
	
	This study described and explained the downside of using sliding windows for data segmentation and feature extraction, followed by a random k-fold cross validation for model evaluation. We identified previous studies from the early years of HAR research until present day publications, where the reported performance is overestimated. The findings, described in detail for each of the reported studies, suggest that this is a recurrent practice with negative effects that are often disregarded by the HAR practitioners.
	
	Our experiments used different types of classification models and they were performed using different datasets in terms of distribution, nature and characteristics, which allows to prove that the bias introduced by the issue discussed in this work is independent of the data modality and the classification model. The performance drop with an unbiased evaluation goes from 5.4\% to 33.6\%. Thus, those highly overestimated models would become impractical or even totally useless in real settings. 
	%Hence, our results show and confirm the drawbacks of windows-based data segmentation followed by random k-fold cross validation.
	
	It is important to clarify that we are not implying that sliding windows and k-fold cross validation should not be used in HAR. Those are well established methods whose usage has been empirically justified. However, the data samples assigned to each fold should not be chosen at random. If data segmentation is performed using a windowing mechanism, the independence between training and test sets, or between folds in CV, must be carefully considered. The group-based approach, used in this work, is a good alternative for guaranteeing an unbiased model evaluation. The ability to split the data by a third-party parameter, chosen per use-case basis, gives the flexibility to create unbiased scenarios for evaluation even for single-subject datasets.

	\bibliographystyle{IEEEtran}
	\bibliography{references}

% Generated by IEEEtran.bst, version: 1.14 (2015/08/26)
\begin{thebibliography}{10}
\providecommand{\url}[1]{#1}
\csname url@samestyle\endcsname
\providecommand{\newblock}{\relax}
\providecommand{\bibinfo}[2]{#2}
\providecommand{\BIBentrySTDinterwordspacing}{\spaceskip=0pt\relax}
\providecommand{\BIBentryALTinterwordstretchfactor}{4}
\providecommand{\BIBentryALTinterwordspacing}{\spaceskip=\fontdimen2\font plus
\BIBentryALTinterwordstretchfactor\fontdimen3\font minus
  \fontdimen4\font\relax}
\providecommand{\BIBforeignlanguage}[2]{{%
\expandafter\ifx\csname l@#1\endcsname\relax
\typeout{** WARNING: IEEEtran.bst: No hyphenation pattern has been}%
\typeout{** loaded for the language `#1'. Using the pattern for}%
\typeout{** the default language instead.}%
\else
\language=\csname l@#1\endcsname
\fi
#2}}
\providecommand{\BIBdecl}{\relax}
\BIBdecl

\bibitem{van2008accurate}
T.~Van~Kasteren, A.~Noulas, G.~Englebienne, and B.~Kr{\"o}se, ``Accurate
  activity recognition in a home setting,'' in \emph{Proceedings of the 10th
  international conference on Ubiquitous computing}, 2008, pp. 1--9.

\bibitem{bulling2014tutorial}
A.~Bulling, U.~Blanke, and B.~Schiele, ``A tutorial on human activity
  recognition using body-worn inertial sensors,'' \emph{ACM Computing Surveys
  (CSUR)}, vol.~46, no.~3, pp. 1--33, 2014.

\bibitem{kwapisz2011activity}
J.~R. Kwapisz, G.~M. Weiss, and S.~A. Moore, ``Activity recognition using cell
  phone accelerometers,'' \emph{ACM SigKDD Explorations Newsletter}, vol.~12,
  no.~2, pp. 74--82, 2011.

\bibitem{chen2015deep}
Y.~Chen and Y.~Xue, ``A deep learning approach to human activity recognition
  based on single accelerometer,'' in \emph{2015 IEEE international conference
  on systems, man, and cybernetics}.\hskip 1em plus 0.5em minus 0.4em\relax
  IEEE, 2015, pp. 1488--1492.

\bibitem{hammerla2016deep}
N.~Y. Hammerla, S.~Halloran, and T.~Pl{\"o}tz, ``Deep, convolutional, and
  recurrent models for human activity recognition using wearables,''
  \emph{arXiv preprint arXiv:1604.08880}, 2016.

\bibitem{ordonez2016deep}
F.~J. Ord{\'o}{\~n}ez and D.~Roggen, ``Deep convolutional and lstm recurrent
  neural networks for multimodal wearable activity recognition,''
  \emph{Sensors}, vol.~16, no.~1, p. 115, 2016.

\bibitem{wan2020deep}
S.~Wan, L.~Qi, X.~Xu, C.~Tong, and Z.~Gu, ``Deep learning models for real-time
  human activity recognition with smartphones,'' \emph{Mobile Networks and
  Applications}, vol.~25, no.~2, pp. 743--755, 2020.

\bibitem{hammerla2015let}
N.~Y. Hammerla and T.~Pl{\"o}tz, ``Let's (not) stick together: pairwise
  similarity biases cross-validation in activity recognition,'' in
  \emph{Proceedings of the 2015 ACM international joint conference on pervasive
  and ubiquitous computing}, 2015, pp. 1041--1051.

\bibitem{dehghani2019subject}
A.~Dehghani, T.~Glatard, and E.~Shihab, ``Subject cross validation in human
  activity recognition,'' \emph{arXiv preprint arXiv:1904.02666}, 2019.

\bibitem{dehghani2019quantitative}
A.~Dehghani, O.~Sarbishei, T.~Glatard, and E.~Shihab, ``A quantitative
  comparison of overlapping and non-overlapping sliding windows for human
  activity recognition using inertial sensors,'' \emph{Sensors}, vol.~19,
  no.~22, p. 5026, 2019.

\bibitem{jordao2018human}
A.~Jordao, A.~C. Nazare~Jr, J.~Sena, and W.~R. Schwartz, ``Human activity
  recognition based on wearable sensor data: A standardization of the
  state-of-the-art,'' \emph{arXiv preprint arXiv:1806.05226}, 2018.

\bibitem{gholamiangonabadi2020deep}
D.~Gholamiangonabadi, N.~Kiselov, and K.~Grolinger, ``Deep neural networks for
  human activity recognition with wearable sensors: Leave-one-subject-out
  cross-validation for model selection,'' \emph{IEEE Access}, vol.~8, pp.
  133\,982--133\,994, 2020.

\bibitem{ahad2020iot}
M.~A.~R. Ahad, A.~D. Antar, and M.~Ahmed, ``Iot sensor-based activity
  recognition,'' \emph{IoT Sensor-based Activity Recognition. Springer}, 2020.

\bibitem{khalifa2017harke}
S.~Khalifa, G.~Lan, M.~Hassan, A.~Seneviratne, and S.~K. Das, ``Harke: Human
  activity recognition from kinetic energy harvesting data in wearable
  devices,'' \emph{IEEE Transactions on Mobile Computing}, vol.~17, no.~6, pp.
  1353--1368, 2017.

\bibitem{micucci2017unimib}
D.~Micucci, M.~Mobilio, and P.~Napoletano, ``Unimib shar: A dataset for human
  activity recognition using acceleration data from smartphones,''
  \emph{Applied Sciences}, vol.~7, no.~10, p. 1101, 2017.

\bibitem{san2018robust}
R.~San-Segundo, H.~Blunck, J.~Moreno-Pimentel, A.~Stisen, and
  M.~Gil-Mart{\'\i}n, ``Robust human activity recognition using smartwatches
  and smartphones,'' \emph{Engineering Applications of Artificial
  Intelligence}, vol.~72, pp. 190--202, 2018.

\bibitem{wang2018eating}
S.~Wang, G.~Zhou, Y.~Ma, L.~Hu, Z.~Chen, Y.~Chen, H.~Zhao, and W.~Jung,
  ``Eating detection and chews counting through sensing mastication muscle
  contraction,'' \emph{Smart Health}, vol.~9, pp. 179--191, 2018.

\bibitem{mutegeki2020cnn}
R.~Mutegeki and D.~S. Han, ``A cnn-lstm approach to human activity
  recognition,'' in \emph{2020 International Conference on Artificial
  Intelligence in Information and Communication (ICAIIC)}.\hskip 1em plus 0.5em
  minus 0.4em\relax IEEE, 2020, pp. 362--366.

\bibitem{ni2020leveraging}
Q.~Ni, Z.~Fan, L.~Zhang, C.~D. Nugent, I.~Cleland, Y.~Zhang, and N.~Zhou,
  ``Leveraging wearable sensors for human daily activity recognition with
  stacked denoising autoencoders,'' \emph{Sensors}, vol.~20, no.~18, p. 5114,
  2020.

\bibitem{gupta2021deep}
S.~Gupta, ``Deep learning based human activity recognition (har) using wearable
  sensor data,'' \emph{International Journal of Information Management Data
  Insights}, vol.~1, no.~2, p. 100046, 2021.

\bibitem{li2021tribogait}
J.~Li, Z.~Wang, Z.~Zhao, Y.~Jin, J.~Yin, S.-L. Huang, and J.~Wang, ``Tribogait:
  A deep learning enabled triboelectric gait sensor system for human activity
  recognition and individual identification,'' in \emph{Adjunct Proceedings of
  the 2021 ACM International Joint Conference on Pervasive and Ubiquitous
  Computing and Proceedings of the 2021 ACM International Symposium on Wearable
  Computers}, 2021, pp. 643--648.

\bibitem{mekruksavanich2021lstm}
S.~Mekruksavanich and A.~Jitpattanakul, ``Lstm networks using smartphone data
  for sensor-based human activity recognition in smart homes,'' \emph{Sensors},
  vol.~21, no.~5, p. 1636, 2021.

\bibitem{bouchabou2021fully}
D.~Bouchabou, S.~M. Nguyen, C.~Lohr, B.~Leduc, and I.~Kanellos, ``Fully
  convolutional network bootstrapped by word encoding and embedding for
  activity recognition in smart homes,'' in \emph{International Workshop on
  Deep Learning for Human Activity Recognition}.\hskip 1em plus 0.5em minus
  0.4em\relax Springer, 2021, pp. 111--125.

\bibitem{ramos2021daily}
R.~G{\'o}mez~Ramos, J.~Duque~Domingo, E.~Zalama, and
  J.~G{\'o}mez-Garc{\'\i}a-Bermejo, ``Daily human activity recognition using
  non-intrusive sensors,'' \emph{Sensors}, vol.~21, no.~16, p. 5270, 2021.

\bibitem{zimbelman2021development}
E.~G. Zimbelman and R.~F. Keefe, ``Development and validation of
  smartwatch-based activity recognition models for rigging crew workers on
  cable logging operations,'' \emph{Plos one}, vol.~16, no.~5, p. e0250624,
  2021.

\bibitem{yan2022deep}
Y.~Yan, T.~Liao, J.~Zhao, J.~Wang, L.~Ma, W.~Lv, J.~Xiong, and L.~Wang, ``Deep
  transfer learning with graph neural network for sensor-based human activity
  recognition,'' \emph{arXiv preprint arXiv:2203.07910}, 2022.

\bibitem{wang2022sensor}
J.~Wang, T.~Zhu, J.~Gan, L.~L. Chen, H.~Ning, and Y.~Wan, ``Sensor data
  augmentation by resampling in contrastive learning for human activity
  recognition,'' \emph{IEEE Sensors Journal}, vol.~22, no.~23, pp.
  22\,994--23\,008, 2022.

\bibitem{huang2022channel}
W.~Huang, L.~Zhang, H.~Wu, F.~Min, and A.~Song, ``Channel-equalization-har: a
  light-weight convolutional neural network for wearable sensor based human
  activity recognition,'' \emph{IEEE Transactions on Mobile Computing}, 2022.

\bibitem{luo2021binarized}
F.~Luo, S.~Khan, Y.~Huang, and K.~Wu, ``Binarized neural network for edge
  intelligence of sensor-based human activity recognition,'' \emph{IEEE
  Transactions on Mobile Computing}, vol.~22, no.~3, pp. 1356--1368, 2023.

\bibitem{wu2023novel}
H.~Wu, Z.~Zhang, X.~Li, K.~Shang, Y.~Han, Z.~Geng, and T.~Pan, ``A novel pedal
  musculoskeletal response based on differential spatio-temporal lstm for human
  activity recognition,'' \emph{Knowledge-Based Systems}, vol. 261, p. 110187,
  2023.

\bibitem{garcia2023new}
D.~Garcia-Gonzalez, D.~Rivero, E.~Fernandez-Blanco, and M.~R. Luaces, ``New
  machine learning approaches for real-life human activity recognition using
  smartphone sensor-based data,'' \emph{Knowledge-Based Systems}, p. 110260,
  2023.

\bibitem{cook2012casas}
D.~J. Cook, A.~S. Crandall, B.~L. Thomas, and N.~C. Krishnan, ``Casas: A smart
  home in a box,'' \emph{Computer}, vol.~46, no.~7, pp. 62--69, 2012.

\bibitem{banos2014mhealthdroid}
O.~Banos, R.~Garcia, J.~A. Holgado-Terriza, M.~Damas, H.~Pomares, I.~Rojas,
  A.~Saez, and C.~Villalonga, ``mhealthdroid: a novel framework for agile
  development of mobile health applications,'' in \emph{International workshop
  on ambient assisted living}.\hskip 1em plus 0.5em minus 0.4em\relax Springer,
  2014, pp. 91--98.

\bibitem{reiss2012intro}
A.~Reiss and D.~Stricker, ``Introducing a new benchmarked dataset for activity
  monitoring,'' in \emph{2012 16th international symposium on wearable
  computers}.\hskip 1em plus 0.5em minus 0.4em\relax IEEE, 2012, pp. 108--109.

\bibitem{krishnan2014activity}
N.~C. Krishnan and D.~J. Cook, ``Activity recognition on streaming sensor
  data,'' \emph{Pervasive and mobile computing}, vol.~10, pp. 138--154, 2014.

\bibitem{quigley2018comparative}
B.~Quigley, M.~Donnelly, G.~Moore, and L.~Galway, ``A comparative analysis of
  windowing approaches in dense sensing environments,'' \emph{Multidisciplinary
  Digital Publishing Institute Proceedings}, vol.~2, no.~19, p. 1245, 2018.

\bibitem{banos2014window}
O.~Banos, J.-M. Galvez, M.~Damas, H.~Pomares, and I.~Rojas, ``Window size
  impact in human activity recognition,'' \emph{Sensors}, vol.~14, no.~4, pp.
  6474--6499, 2014.

\bibitem{domingos2012few}
P.~Domingos, ``A few useful things to know about machine learning,''
  \emph{Communications of the ACM}, vol.~55, no.~10, pp. 78--87, 2012.

\bibitem{altun2010comparative}
K.~Altun, B.~Barshan, and O.~Tun{\c{c}}el, ``Comparative study on classifying
  human activities with miniature inertial and magnetic sensors,''
  \emph{Pattern Recognition}, vol.~43, no.~10, pp. 3605--3620, 2010.

\bibitem{stisen2015smart}
A.~Stisen, H.~Blunck, S.~Bhattacharya, T.~S. Prentow, M.~B. Kj{\ae}rgaard,
  A.~Dey, T.~Sonne, and M.~M. Jensen, ``Smart devices are different: Assessing
  and mitigatingmobile sensing heterogeneities for activity recognition,'' in
  \emph{Proceedings of the 13th ACM conference on embedded networked sensor
  systems}, 2015, pp. 127--140.

\bibitem{anguita2013public}
D.~Anguita, A.~Ghio, L.~Oneto, X.~Parra~Perez, and J.~L. Reyes~Ortiz, ``A
  public domain dataset for human activity recognition using smartphones,'' in
  \emph{Proceedings of the 21th international European symposium on artificial
  neural networks, computational intelligence and machine learning}, 2013, pp.
  437--442.

\bibitem{casale2011human}
P.~Casale, O.~Pujol, and P.~Radeva, ``Human activity recognition from
  accelerometer data using a wearable device,'' in \emph{Iberian conference on
  pattern recognition and image analysis}.\hskip 1em plus 0.5em minus
  0.4em\relax Springer, 2011, pp. 289--296.

\bibitem{weiss2019smartphone}
G.~M. Weiss, K.~Yoneda, and T.~Hayajneh, ``Smartphone and smartwatch-based
  biometrics using activities of daily living,'' \emph{IEEE Access}, vol.~7,
  pp. 133\,190--133\,202, 2019.

\bibitem{malekzadeh2018protecting}
M.~Malekzadeh, R.~G. Clegg, A.~Cavallaro, and H.~Haddadi, ``Protecting sensory
  data against sensitive inferences,'' in \emph{Proceedings of the 1st Workshop
  on Privacy by Design in Distributed Systems}, 2018, pp. 1--6.

\bibitem{zhang2012usc}
M.~Zhang and A.~A. Sawchuk, ``Usc-had: A daily activity dataset for ubiquitous
  activity recognition using wearable sensors,'' in \emph{Proceedings of the
  2012 ACM conference on ubiquitous computing}, 2012, pp. 1036--1043.

\bibitem{chatzaki2017human}
C.~Chatzaki, M.~Pediaditis, G.~Vavoulas, and M.~Tsiknakis, ``Human daily
  activity and fall recognition using a smartphone’s acceleration sensor,''
  in \emph{Information and Communication Technologies for Ageing Well and
  e-Health: Second International Conference, ICT4AWE 2016, Rome, Italy, April
  21-22, 2016, Revised Selected Papers 2}.\hskip 1em plus 0.5em minus
  0.4em\relax Springer, 2017, pp. 100--118.

\bibitem{roggen2010collecting}
D.~Roggen, A.~Calatroni, M.~Rossi, T.~Holleczek, K.~F{\"o}rster,
  G.~Tr{\"o}ster, P.~Lukowicz, D.~Bannach, G.~Pirkl, A.~Ferscha \emph{et~al.},
  ``Collecting complex activity datasets in highly rich networked sensor
  environments,'' in \emph{2010 Seventh international conference on networked
  sensing systems (INSS)}.\hskip 1em plus 0.5em minus 0.4em\relax IEEE, 2010,
  pp. 233--240.

\bibitem{luo2019kitchen}
F.~Luo, S.~Poslad, and E.~Bodanese, ``Kitchen activity detection for healthcare
  using a low-power radar-enabled sensor network,'' in \emph{ICC 2019-2019 IEEE
  International Conference on Communications (ICC)}.\hskip 1em plus 0.5em minus
  0.4em\relax IEEE, 2019, pp. 1--7.

\bibitem{cook2009assessing}
D.~J. Cook and M.~Schmitter-Edgecombe, ``Assessing the quality of activities in
  a smart environment,'' \emph{Methods of information in medicine}, vol.~48,
  no.~05, pp. 480--485, 2009.

\bibitem{wang2021multi}
T.~Wang and D.~J. Cook, ``Multi-person activity recognition in continuously
  monitored smart homes,'' \emph{IEEE Transactions on Emerging Topics in
  Computing}, 2021.

\bibitem{anguita2013energy}
D.~Anguita, A.~Ghio, L.~Oneto, F.~X. Llanas~Parra, and J.~L. Reyes~Ortiz,
  ``Energy efficient smartphone-based activity recognition using fixed-point
  arithmetic,'' \emph{Journal of universal computer science}, vol.~19, no.~9,
  pp. 1295--1314, 2013.

\bibitem{scikit-learn}
F.~Pedregosa, G.~Varoquaux, A.~Gramfort, V.~Michel, B.~Thirion, O.~Grisel,
  M.~Blondel, P.~Prettenhofer, R.~Weiss, V.~Dubourg, J.~Vanderplas, A.~Passos,
  D.~Cournapeau, M.~Brucher, M.~Perrot, and E.~Duchesnay, ``Scikit-learn:
  Machine learning in {P}ython,'' \emph{Journal of Machine Learning Research},
  vol.~12, pp. 2825--2830, 2011.

\bibitem{Morris:2019weisfeiler}
C.~Morris, M.~Ritzert, M.~Fey, W.~L. Hamilton, J.~E. Lenssen, G.~Rattan, and
  M.~Grohe, ``Weisfeiler and leman go neural: Higher-order graph neural
  networks,'' in \emph{Proceedings of the AAAI conference on artificial
  intelligence}, vol.~33, no.~01, 2019, pp. 4602--4609.

\end{thebibliography}
	
\end{document}